\documentclass[runningheads]{llncs}

\usepackage{eccv}

\usepackage{eccvabbrv}

\usepackage{graphicx}
\usepackage{booktabs}

\usepackage[accsupp]{axessibility}  

\usepackage{tikz}
\usepackage{multirow}
\usepackage{graphicx}
\newcommand{\repeatthanks}{\textsuperscript{\thefootnote}}

\usepackage{hyperref}
\usepackage[misc]{ifsym}
\usepackage{orcidlink}

\usepackage{color,colortbl}
\definecolor{Gray}{gray}{0.92}
\newcolumntype{a}{>{\columncolor{Gray}}c}

\begin{document}

\title{HGL: Hierarchical Geometry Learning for Test-time Adaptation in 3D Point Cloud Segmentation} 

\titlerunning{HGL: Test-time Adaptation in Point Cloud Segmentation}

\author{Tianpei Zou\inst{1}\thanks{Equal Contribution}\ \and Sanqing Qu \inst{1}\repeatthanks \and Zhijun Li \inst{1}\and \\
Alois Knoll \inst{2} \and Lianghua He \inst{1} \and Guang Chen\inst{1}$^{(\textrm{\Letter})}$ \and Changjun Jiang\inst{1} }

\authorrunning{T.~Zou and S.~Qu et al.}


\institute{Tongji University, \and Technical University of Munich \\ \email{\{2011459, 2011444, guangchen\}@tongji.edu.cn}}

\maketitle

\begin{abstract}
    3D point cloud segmentation has received significant interest for its growing applications. However, the generalization ability of models suffers in dynamic scenarios due to the distribution shift between test and training data. To promote robustness and adaptability across diverse scenarios, test-time adaptation (TTA) has recently been introduced. 
    Nevertheless, most existing TTA methods are developed for images, and limited approaches applicable to point clouds ignore the inherent hierarchical geometric structures in point cloud streams, i.e., local (point-level), global (object-level), and temporal (frame-level) structures.
    In this paper, we delve into TTA in 3D point cloud segmentation and propose a novel Hierarchical Geometry Learning (HGL) framework.
    HGL comprises three complementary modules from local, global to temporal learning in a bottom-up manner.
    Technically, we first construct a local geometry learning module for pseudo-label generation. Next, we build prototypes from the global geometry perspective for pseudo-label fine-tuning. Furthermore, we introduce a temporal consistency regularization module to mitigate negative transfer. Extensive experiments on four datasets demonstrate the effectiveness and superiority of our HGL. Remarkably, on the SynLiDAR to SemanticKITTI task, HGL achieves an overall mIoU of 46.91\%, improving GIPSO by 3.0\% and significantly reducing the required adaptation time by 80\%. The code is available at \href{https://github.com/tpzou/HGL}{https://github.com/tpzou/HGL}.
\end{abstract}

\section{Introduction}
\label{sec:intro}

\par 3D point cloud segmentation is essential for many real-world applications, such as medical robotics~\cite{balsiger2019learning, liu2023grab}, geo-surveying~\cite{mugnai2021laser} and autonomous driving~\cite{wu2018squeezeseg, choy20194d, he2023density}. Along with the rapid development of deep neural networks (DNNs)~\cite{he2016deep, dosovitskiy2020image, qi2017pointnet}, it has received significant interest and achieved impressive performance in the past decade. However, this success hinges on the prerequisite that both the training and test data originate from the same distribution. When deploying models to point clouds captured from different sensor configurations or dissimilar scene distributions, inevitable performance degeneration commonly occurs~\cite{yi2021complete, wu2019squeezesegv2}. 

\par To address this challenge, several appealing solutions have been developed, including domain generalization (DG)~\cite{sanchez2023domain, kim2023single, qu2022bmd, qu2023modality} and domain adaptation (DA)~\cite{qin2019pointdan,langer2020domain, wu2019squeezesegv2, yi2021complete, zhao2021epointda}. Specifically, DG aims to learn a model using data from multiple diverse source domains that can generalize well to distribution-shifted target domains. DA utilizes the transductive learning manner to realize knowledge transfer from labeled source domains to unlabeled target domains. Despite showing encouraging results, their real-world applications are still restricted. In particular, DG primarily operates during the training phase and cannot fully utilize test data from application scenarios. DA typically necessitates access to both labeled source data and test data concurrently, constituting offline learning and can be hindered by privacy concerns or legal constraints. In many practical scenarios, DNNs require continual and adaptive updating in an online fashion to remain effective. Hence, we focus on an extreme case in domain adaptation: test-time adaptation (TTA), in which model adaptation is performed by learning from the unlabeled test (target) data during inference time~\cite{wang2020tent}. 

\par Currently, most TTA approaches are proposed for 2D image data. The main techniques include batch normalization calibration~\cite{mirza2022norm}, entropy minimization~\cite{niu2022efficient, wang2020tent}, pseudo-labeling~\cite{chen2022contrastive, goyal2022test}, etc. However, the direct application of these techniques, originally devised for 2D images, onto 3D point cloud data is not practical and yields negligible results~\cite{shin2022mm}. The challenge stems from the intrinsic property of 3D point cloud data, characterized by unordered geometric structures, and the distribution shift is primarily attributed to disparities in the geometric structural information between the source and target domains. 
More recently, GIPSO~\cite{saltori2022gipso}, designed specifically for TTA in 3D point cloud segmentation (TTA-3DSeg), leverages geometric and temporal information to mitigate the domain shift. Nevertheless, GIPSO simply employs geometric features for pseudo-label propagation with a frozen source model and lacks a thorough exploration of hierarchical geometry information, \textcolor{black}{which prevents the pseudo-labels from benefiting by increasing target data}. Furthermore, GIPSO leverages an additional model to extract geometric information, significantly impacting efficiency.

\par In this work, we target TTA-3DSeg from the viewpoint of parsing continuous unstructured point cloud streams into local (point-level), global (object-level), and temporal (frame-level) hierarchical structures for accurate and robust self-supervised learning. The motivation is threefold: at the point level, each individual point exhibits semantic attributes highly correlated with its neighboring points; at the object level, despite distribution shifts, objects of the same category should manifest similar feature embeddings; and at the frame level, the embedding features of geometrically corresponding points from different frames should demonstrate similarity. To materialize our idea, we propose a novel Hierarchical Geometry Learning (HGL) framework.  HGL comprises three complementary modules from local, global to temporal learning in a bottom-up manner.
Technically, we first generate pseudo-labels from the local point level by employing K-NN information aggregation. Then, we build prototypes from the global object level for pseudo-label fine-tuning {with increasing target data}. Thereafter, at the frame level, we introduce temporal geometry consistency regularisation to mitigate negative transfer. We validate the superiority of our HGL via extensive experiments on four benchmarks including various distribution-shift scenarios, e.g., synthetic-to-real, real-to-real, continual learning, and adverse weather. Empirical results show that HGL yields new state-of-the-art performance. It is important to note that HGL is a general framework, the specific implementations within each module are not firmly fixed and can be updated to incorporate advancements in the field. This adaptability is further corroborated by preliminary results presented in Table~{\ref{table:novelty}}.

\par Our contributions can be summarized as follows:
\begin{itemize}
    \item We argue that parsing continuous point cloud streams into point-, object-, and frame-level contributes significantly to the exploration and utilization of geometric information within the unstructured point cloud.
    \item We propose a \textcolor{black}{systematic and expandable} Hierarchical Geometry Learning (HGL) framework, which comprises three components: a local geometry learning module for pseudo-label generation, a global geometry fine-tuning module for pseudo-label fine-tuning, and a temporal geometry regularization module for consistency regularization.
    \item The incorporation of three-level geometry learning enables HGL to achieve high adaptability and efficiency. Extensive experiments on synthetic-to-real, real-to-real, continual learning, and adverse weather support the superiority. Compared to GIPSO~\cite{saltori2022gipso}, HGL achieves 3.0\% improvement in mIoU and 80\% reduction in adaptation time on SynLiDAR to SemanticKITTI task. 
\end{itemize}

\section{Related Work}


\noindent \textbf{Domain Generalization for Point Cloud Segmentation:}  Domain generalization (DG) aims to generalize the model to distribution-shifted target domains by learning domain-invariant features at the training stage. 
For point cloud segmentation,  \cite{kim2023single} augments the source domain to simulate the unseen target domains by randomly subsampling the LiDAR scans. 3DLabelProp~\cite{sanchez2023domain} relies on exploiting the geometry and sequence of LiDAR data to improve its generalization performance by partially accumulated point clouds. LiDOG~\cite{saltori2023walking} introduces an additional BEV auxiliary task to learn the robust features. Although DG can generalize to multiple domains, its inability to leverage data from application scenarios results in sub-optimal performance.

\noindent \textbf{Domain Adaptation for Point Cloud Segmentation:} Domain adaptation (DA) aims to decrease the performance gap of the model between the labeled source domain and the unlabeled target domain. To achieve this, Complete \& Label~\cite{yi2021complete} addresses the differences between point clouds with different LiDAR configurations by completing both domains onto a 3D surface before segmentation. ePointDA~\cite{zhao2021epointda} achieves domain transfer by using a Generative Adversarial Network to simulate real-world noise in virtual data. LiDAR-UDA~\cite{shaban2023lidar} introduces pseudo-label enhancement techniques including LiDAR beam subsampling and cross-frame ensembling for self-supervised training. Despite the success, concurrent access to source and target data still hinders real-world applications.

\noindent \textbf{Test-time Adaptation:} To mitigate the limitations of DG and DA, test-time adaptation (TTA) has been proposed, which strives to adapt models in an online learning manner during inference. TTA has been widely studied in the image domain including recognition ~\cite{wang2020tent, niu2022efficient, chen2022contrastive, qu2023upcycling, qu2024lead}, detection~\cite{kim2022ev, vs2023towards, veksler2023test}, and segmentation~\cite{wang2023dynamically, zhang2022auxadapt}. 
Compared to the well-established image-specific TTA, point-specific TTA focuses primarily on registration\cite{hatem2023point} and upsampling\cite{hatem2023test}, while TTA-3DSeg is still an open research problem. Recently, GIPSO~\cite{saltori2022gipso} utilizes geometric features for pseudo-label propagation and constrains the temporal consistency between consecutive frames. Nonetheless, GIPSO lacks a thorough exploration of hierarchical geometry and requires additional networks to extract geometric information. In this paper, we introduce a novel Hierarchical Geometry Learning (HGL) framework for effective and efficient TTA-3DSeg.

\begin{figure*}[tbp]
    \centering
 \includegraphics[width=0.95\textwidth]{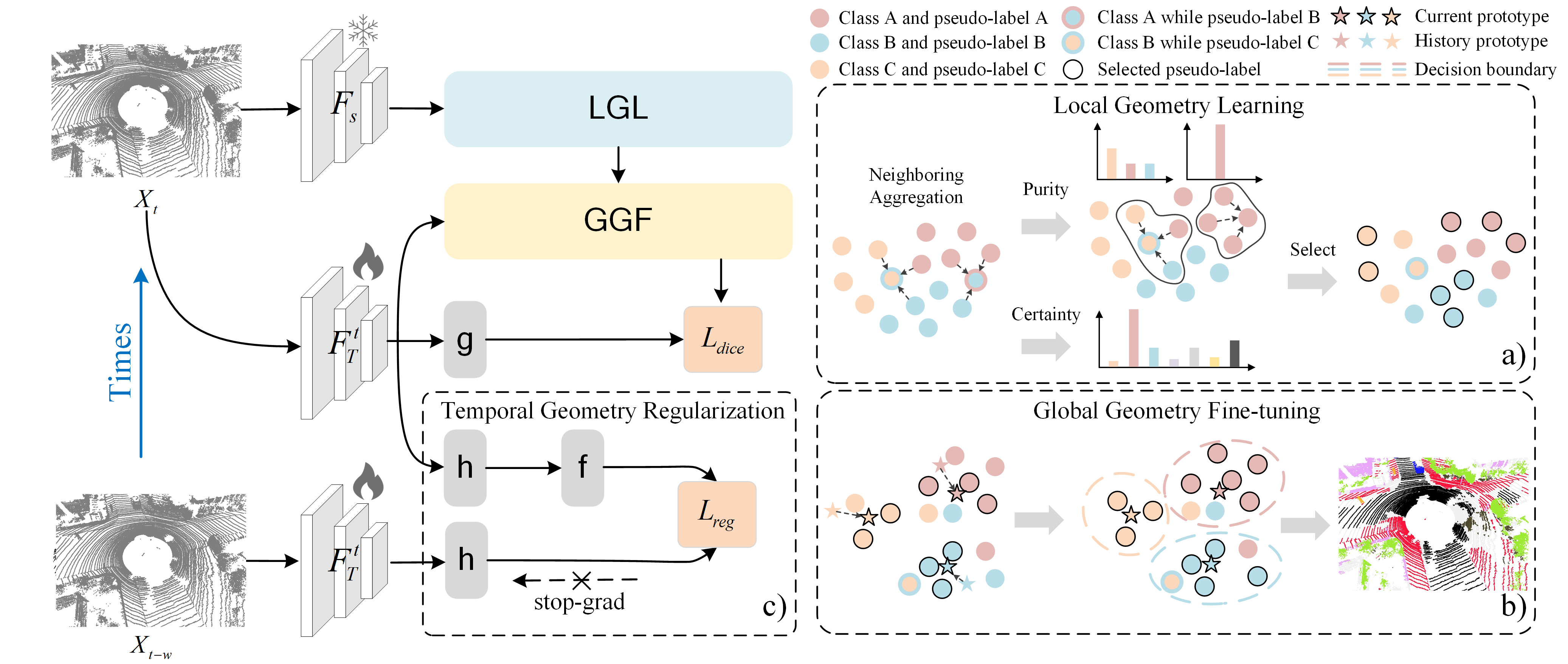}
    \caption{Overview of Hierarchical Geometry Learning (HGL) framework. HGL is composed of a local geometry learning (LGL) module, a global geometry fine-tuning (GGF) module, and a temporal geometry regularization module.  Specifically, a) LGL exploits local K-NN clustering strategy to generate local geometric pseudo-labels based on neighboring aggregation, prediction certainty, and geometric purity. Then, b) GGF builds the prototype to explore global geometry for pseudo-label fine-tuning.  Besides, c) encourages the features of geometrically corresponding points between different frames to remain consistent, where $h( \cdot )$ means encoder network and $f( \cdot )$ means predictor head.}
    \label{fig:fig_framework}
\end{figure*}

\section{Methodology}
\subsection{Definitions}
\par In this paper, we focus on TTA-3DSeg, allowing a pre-trained model $F_S$ to be updated online to a new target domain $T$ without access to the source domain $S$. Formally, the data of the source domain are defined as $\mathcal{D}_s = \{(X^i_S, Y^i_S)\}^{N_S}_{i=1}$ where $X^i_S$ is labeled point cloud, $Y^i_S$ is corresponding segmentation label of $X^i_S$ and $N_S$ is the number of labeled point clouds. $F_T$ is the target model adapted from the pre-trained model $F_S$. The data of the target domain are denoted as $\mathcal{D}_t = \{(X_T^t, Y_T^t)\}^{N_T}_{t=1}$ where $X^i_T$ represents an unlabeled point cloud of the stream at time $t$ and $Y_T^t$ is unknown label. The source and target classes are coincident.

\subsection{Our Approach}
\par To adapt source model $F_S$ to target model $F_T$, we propose a Hierarchical Geometry Learning (HGL) framework for accurate and robust self-supervised learning from local (point-level), global (object-level) and temporal (frame-level) hierarchical structures. As shown in Fig.~\ref{fig:fig_framework}, our HGL includes three complementary components, the local geometry learning module, the global geometry fine-tuning module, and the temporal geometry regularization module. 

\par Technically, given the current point cloud $X_T^t$, we first input it into the local geometry learning module to generate local geometric pseudo-labels and select reliable samples. The local geometry learning module is based on the local K-NN clustering strategy, which includes aggregating predictions for each point from its nearest neighbors and calculating confidence by both geometric purity and prediction certainty. The geometric purity represents the purity score of surrounding points, obtained by calculating anti-entropy based on the predictions of neighboring points. 
Besides, the prediction certainty is the anti-entropy of each point prediction. 
Based on selected local geometric pseudo-labels, we further introduce a global geometry fine-tuning module, in which we build prototypes for each class object on the target model and employ them as a new classifier to obtain global geometric pseudo-labels. By incorporating local and global geometric pseudo-labels, we could obtain robust and generalized local-global geometric pseudo-labels.
Lastly, we introduce a temporal geometry regularization module to constrain the consistency of the embedding feature of geometrically corresponding points between the current point cloud $X_T^t$ and the previous $X_T^{t-w}$. In contrast to forcing regularization at all points~\cite{saltori2022gipso}, we align low-confidence to high-confidence features to avoid degradation of high-confidence features.

\subsection{Local Geometry Learning}

\par As a promising technique in unsupervised learning, pseudo-labeling has been widely employed. Existing methods~\cite{shin2022mm, saltori2022gipso,wang2023space} typically generate pseudo-labels directly based on single-point prediction. Despite reasonable effectiveness, these methods neglect the local geometry structure, leading to sub-optimal results.
To tackle this, as shown in Fig.~\ref{fig:fig_framework} a), we propose the Local Geometric Learning (LGL) module to fully leverage local point-level information. We first aggregate the predictions from spatial neighbors to generate more accurate pseudo-labels. Then, we introduce geometric purity and prediction certainty confidence to select reliable local geometric pseudo-labels.
The local geometric learning design stems from an intuitive observation that spatially adjacent points often contain the same semantics due to the inherent geometric structure in point clouds.

\par \textbf{Neighboring aggregation.} Given a point cloud $X_T^t$, we first pass it through the frozen source model $F_S$, and obtain the softmax output, i.e., prediction $P_T^t \in R^{N \times C}$. Then, for each point $x_i^t$ in $X_T^t$, we leverage K-NN to locate its geometric neighbors, and subsequently obtain aggregated prediction $\hat{p}_i$ by weighted-summing the neighbors' prediction with respect to the distance. Specifically,
\begin{equation}
\begin{aligned}
     & \hat{p}_i^t = \frac{1}{\sum_{j\in \mathcal{N}_i} w_j^t}\sum_{j\in \mathcal{N}_i} w_j^t p_j^t \\
     &  w^t_j = e^{-\mathrm{dis}(x^t_i, x^t_j)}
\end{aligned} 
\end{equation}
where $w_j^t$ represents the weight and $\mathrm{dis}(,)$ denotes the Euclidean distance. It is worth noting that $\mathcal{N}_i$ is the $K+1$ nearest neighbors of $x_i^t$, as we include the point itself as one of the neighbors. This ensures that self-prediction is not overlooked when incorporating with predictions from surrounding points. The local geometric pseudo-labels $\hat{y}_i^{t,local}$ are produced as 
\begin{equation}
\hat{{y}}_i^{t,local}=\mathbf{1}_c\left(\underset{c}{\arg \max }\left\{\hat{{p}}_{i, c}^t\right\}_{c \in C}\right)
\end{equation}
where $\mathbf{1}_c(.)$ denotes the one-hot transformation.

\par \textbf{Prediction certainty.} Calculating the confidence of point predictions and using this information for reliable pseudo-label selection is crucial to self-supervised learning. To achieve this, we measure the prediction certainty $\mathcal{C}_i^t$ by employing the normalized Shannon Entropy~\cite{shannon1948mathematical} as:
\begin{equation}
    \mathcal{C}_i^t = 1 - \frac{1}{\log C} \sum_{c=1}^{C} \hat{p}^t_{i,c} \log \hat{p}^t_{i,c}
\end{equation}
where $C$ is the class number, and $\hat{p}_{i,c}^t$ denotes the soft-max probability of point $x_i$ belonging to the $c$-th class.

\par \textbf{Geometric purity.} In addition to measuring certainty solely based on predictions, we further introduce a geometric purity metric to locate reliable local regions by measuring the consistency of predictions in local regions.
Specifically, given the aggregation prediction $\hat{y}_i^{t,local}$ and neighbors $\mathcal{N}_i$, we consider it as a local geometric region and calculate geometric purity based on the point cloud class distribution within the region. Formally, we normalize the $\hat{y}_j^{t,local}$ within the local geometric region to obtain probability $\mathcal{P}^t$, and then calculate geometric purity $\mathcal{A}$ with reference to normalized Shannon Entropy, which is formulated as:
\begin{equation}
\begin{aligned}
    \mathcal{P}_i^t &= \frac{1}{\mathcal{N}_i} \sum_{j\in \mathcal{N}_i} (\hat{y}_j^{t,local}) \\
    \mathcal{A}_i^t &= 1 - \frac{1}{\log C} \sum_{c=1}^{C} \mathcal{P}_{i,c}^t \log \mathcal{P}_{i,c}^t 
\end{aligned} 
\end{equation}

\par Accordingly, we obtain the final function of confidence score as $\mathcal{S} = \mathcal{C} \cdot \mathcal{A}$. Then, we group $\mathcal{S}$ scores per class and determine  $a$ as the $\lambda$ -th percentile of $\mathcal{S}$ for class $c$. Consequently, at time $t$ and for class $c$, we exclusively choose local geometric pseudo-labels with confidence scores higher than $a$, utilizing them as selected local geometric pseudo-labels. Note that LGL is based on the frozen source model, rather than the EMA-based update model. 
The reason behind this is that despite generally better performance in downstream tasks, EMA may still put the model at risk of catastrophic forgetting and representational collapse by pushing the model far away from the pre-trained one~\cite{shu2023clipood}.
Relevant experiments can be found in the supplementary materials.

\subsection{Global Geometry Fine-tuning }
\par Although the pseudo-labels have been enhanced with local geometry through LGL, there are some limitations. 
Firstly, the local geometric pseudo-labels are obtained from the source model only. Therefore, the generated pseudo-labels are still noisy due to domain shift and cannot benefit from the continuously increasing target domain data. 
Secondly, the selected local geometric pseudo-labels are chosen by $\lambda$. It is challenging to ensure a good balance between accuracy and generalization in the selected samples. 
To address these limitations and inspired by~\cite{zhang2021prototypical}, we propose the global geometry fine-tuning (GGF) module as shown in Fig.~\ref{fig:fig_framework} b), which fine-tunes local geometric pseudo-labels by calculating the global affinity between the global target prototypes and the feature of each point. 

\par Specifically, to obtain global object information, we first build a set of prototypes for each class on the learnable target model by using selected local geometric pseudo-labels. Then, we obtain the global pseudo-labels by calculating the global affinity and retrieving the nearest prototype for each point. We calculate the centroid of all selected points embedding feature in the class $c$ as the prototype $\rho_c$, which can be formulated as: $\rho_c^t=\frac{1}{\mathcal{N}_c^t} \sum_i^{\mathcal{N}_c^t} z_i^t$, where $\mathcal{N}_c^t$ is the total number of points of current class $c$ in selected local geometric pseudo-labels and $z_i^t$ is the embedding feature of the point in the target model. 
Different from local geometric pseudo-labels obtained from the frozen source model, we update the prototype along the sequential iterations using EMA as follows: $\hat{\rho}_c^t=\alpha \hat{\rho}_c^{t-1}+(1-\alpha)\rho_c^t$
where $\hat{\rho}_c^0={\rho}_c^0$, $\hat{{\rho}}_c^t$ and $\hat{{\rho}}_c^{t-1}$ mean the current $t$-th prototype and last $(t-1)$-th prototype in iterations, and $\alpha$ is a hyper-parameter that controls the updating speed. Given the prototype $\hat{{\rho}}_c^t$ and point embedding feature $z_i^t$, the global geometric pseudo-label is attained by:
\begin{equation}
\hat{y}_i^{t, global}=\mathbf{1}_{{c}}\left(\underset{c}{\arg \max }\left\{\mathrm{sim}\left(z_i^{t}, \hat{\rho}_c^t\right)\right\}_{c \in C}\right),
\end{equation}
where $\mathrm{sim}(a, b)$ measures the similarity between $a$ and $b$. We apply the cosine similarity function by default.
With the global geometric pseudo-labels, we propose a straightforward strategy to fine-tune the local geometric pseudo-labels to obtain more reliable local-global pseudo-labels, i.e., retaining only the pseudo-labels in the local that are consistent with the global:
\begin{equation}
\label{final_label}
\hat{y}_i^{t} = \hat{y}_i^{t, local}, \mathrm{s.t.} \ \hat{y}_i^{t, global} = \hat{y}_i^{t, local}
\end{equation}
It is worth noting that to ensure the generalization, we operate on all instances of local geometric pseudo-labels here, rather than just the selected subset. Through the optimization with global geometric pseudo-labels, we introduce continuously optimized temporal information while ensuring the non-collapse of pseudo-labels. Moreover, we achieve a balance between accuracy and generalization through a local-global pseudo-label filtering process.

\subsection{Temporal Geometry Regularization }
Unlike TTA in image recognition~\cite{wang2020tent}, another significant characteristic of TTA-3DSeg is the temporal geometry consistency. Considering the temporal geometric consistency in local-global pseudo-labeling-based self-supervised learning is effective in smoothing the prediction and preventing negative transfer.

\par Follow~\cite{chen2021exploring, saltori2022gipso}, we begin by computing the geometry matching point pairs between $X_T^{t-w}$ and $X_T^t$ by using the rigid transformation $T^{t-w \to t}$. Subsequently, as shown in Fig.~\ref{fig:fig_framework} c), we add an encoder network $h( \cdot )$ and a predictor head $f( \cdot )$ to project point cloud pairs into a representation space to obtain corresponding features $q, z$ and minimize the negative cosine similarity between them.
\begin{equation}
    \label{tcc}
    \mathcal{D}_{t \rightarrow t-w}\left(q^t, z^{t-w}\right)=-\frac{q^t}{\left\|q^t\right\|_2} \cdot \frac{z^{t-w}}{\left\|z^{t-w}\right\|_2} 
\end{equation}
Different from~\cite{chen2021exploring, saltori2022gipso}, we argue that due to the typically higher accuracy of high-confidence predictions, it is reasonable to further demand the alignment of low-confidence features with their high-confidence counterparts, which effectively prevents the corrupting of high-confidence features by low-confidence ones. Thus, we make modification to Eq. \ref{tcc} by introducing the geometry confidence score $\mathcal{S}$:
\begin{equation}
    \mathcal{D}^{\prime}_{t \rightarrow t-w}\left(q^t, z^{t-w}\right)=-\mathcal{S}^{t-w}\frac{q^t}{\left\|q^t\right\|_2} \cdot \frac{z^{t-w}}{\left\|z^{t-w}\right\|_2} 
\end{equation}
The final temporal consistency loss is defined as:
\begin{equation}
    \label{reg_loss}
    \mathcal{L}_{reg}=\frac{1}{2} \mathcal{D}^{\prime}_{t \rightarrow t-w}\left(q^t, z^{t-w}\right)+\frac{1}{2} \mathcal{D}^{\prime}_{t-w \rightarrow t}\left(q^{t-w}, z^t\right)
\end{equation}
Please refer to supplementary materials for more details.

\subsection{Online model update}
\par In point clouds, class imbalances are typically pronounced. Furthermore, our local-global pseudo-label generation process may still introduce some erroneous low-confidence pseudo-labels. To this end, we employ the soft Dice loss $\mathcal{L}_{dice}$~\cite{jadon2020survey}, using smooth labels~\cite{liang2020we} and the pseudo-labels selected through Eq.\ref{final_label} as supervision, where the smoothing parameter is controlled by the confidence score $\mathcal{S}$. More detailed descriptions can be found in appendix. The overall objective is $\mathcal{L}_{final} = \mathcal{L}_{dice} + \mathcal{L}_{reg}$, where $\mathcal{L}_{reg}$ is temporal consistency loss defined in Eq.\ref{reg_loss}.

\section{Experiments}
\subsection{Setup}
\noindent \textbf{Dataset:} We empirically verify the effectiveness and versatility of HGL on two virtual datasets (SynLiDAR~\cite{xiao2021synlidar} and Synth4D~\cite{saltori2022gipso}) and two real-world datasets (SemanticKITTI~\cite{behley2019semantickitti} and nuScenes~\cite{caesar2020nuscenes}).
\textbf{SynLiDAR}~\cite{xiao2021synlidar} contains 98,396 frames of point clouds synthesized by the Unreal Engine 4 platform, which can be used to simulate SemantiKITTI. \textbf{Synth4D}~\cite{saltori2022gipso} contains 20K LiDAR point clouds synthesized by the CARLA simulator~\cite{dosovitskiy2017carla}, which can be used for simulating SemantiKITTI and nuScenes.
\textbf{SemanticKITTI}~\cite{behley2019semantickitti} is a large-scale real-world dataset for LiDAR point-cloud segmentation with 21 sequences and 43,442 densely annotated scans at 10Hz. We follow the official sequence split and use 00-07, 09-10 scenes as source domain data (19,130 frames) and scene 08 as target domain data (4,071 frames). \textbf{nuScenes}~\cite{caesar2020nuscenes} contains 40K LiDAR frames annotated with per-point semantic labels from 1K sequences at 2Hz, each with approximately 40 scans in each sequence.  
We employ 700 sequences as source domain data (28,130 frames) and 150 sequences as target domain data (6,019 frames). 
Due to the different annotations across datasets, we remapped semantic categories into seven standardized classes to ensure consistent category definitions. Please refer to the supplementary materials for more information.

\begin{table*}[tbp]
\centering
\caption{mIoU improvement (\%) comparison in SynLiDAR to SemanticKITTI test-time adaptation. $*$ denotes being re-purposed for test-time adaptation. }
\addtolength{\tabcolsep}{+2.0pt}
\resizebox{0.99\textwidth}{!}{
\begin{tabular}{lccccccca}
\toprule
\textbf{Model} & \textbf{Vehicle} & \textbf{Pedestrian} & \textbf{Road} & \textbf{Sidewalk} & \textbf{Terrain} & \textbf{Manmade} & \textbf{Vegetation} & \textbf{Avg} \\
\midrule
Source         & 59.80            & 14.20               & 34.90         & 53.50             & 31.00            & 37.40            & 50.50               & 40.19        \\
Upper Bound         & +21.32           & +8.09               & +11.51        & +28.13            & +40.46           & +33.67           & +30.63              & +24.83       \\
\midrule
ADABN~\cite{li2016revisiting}          & +3.90            & -6.40               & -0.20         & -3.70             & -5.70            & +1.40            & +0.30               & -1.49        \\
ProDA*~\cite{zhang2021prototypical}          & -53.30           & -13.79              & -33.83        & -52.78            & -30.52           & -36.68           & -49.29              & -38.60       \\
SHOT*~\cite{liang2020we}           & -57.83           & -12.64              & -24.80        & -46.02            & -30.80           & -36.83           & -49.32              & -36.89       \\
CBST*~\cite{zou2018unsupervised}           & +0.99            & -0.83               & +0.55         & +0.20             & +0.74            & -0.07            & +0.38               & +0.28        \\

TENT~\cite{wang2020tent}          &-0.27            &-3.54              &+1.63         &+1.49              &-0.33             &+4.96            &+4.15              & +1.15  \\
ConjugatePL~\cite{goyal2022test}          &+4.16            &-0.73              &+1.82         &+1.80              &-1.36             &+5.27            &+4.95              & +2.27 \\

GIPSO~\cite{saltori2022gipso}          & +13.95           & -6.76               & \textbf{+3.26}         & +5.01             & +3.00            & +3.34            & +4.08               & +3.70        \\
\midrule
HGL(Ours)     & \textbf{+14.76}           & \textbf{+5.66}               & +1.83         & \textbf{+5.43}             & \textbf{+7.33}            & \textbf{+5.64}            & \textbf{+6.40}              & \textbf{+6.72}       \\
\bottomrule
\end{tabular}
}
\label{table:syn2kitti}
\end{table*}

\begin{table*}[tbp]
\centering
\caption{mIoU improvement (\%) comparison in Synth4D to SemanticKITTI test-time adaptation. $*$ denotes being re-purposed for test-time adaptation. }
\addtolength{\tabcolsep}{+2.0pt}
\resizebox{0.99\textwidth}{!}{
\begin{tabular}{lccccccca}
\toprule
\textbf{Model} & \textbf{Vehicle} & \textbf{Pedestrian} & \textbf{Road} & \textbf{Sidewalk} & \textbf{Terrain} & \textbf{Manmade} & \textbf{Vegetation} & \textbf{Avg} \\
\midrule
Source         & 63.90            & 12.60               & 38.10          & 47.30             & 20.20            & 26.10            & 43.30               & 35.93          \\
Upper Bound         & +16.84           & +5.49               & +8.48          & +34.44            & +51.92           & +45.68           & +39.09              & +28.85         \\
\midrule
ADABN~\cite{li2016revisiting}          & -7.80            & -2.00               & -10.20         & -18.60            & -7.70            & +5.80            & -0.70               & -5.89          \\
ProDA*~\cite{zhang2021prototypical}          & -57.77           & -12.34              & -37.36         & -46.95            & -19.97           & -25.62           & -42.48              & -34.64         \\
SHOT*~\cite{liang2020we}           & -62.44           & -12.00              & -28.27         & -40.20            & -20.00           & -25.47           & -42.55              & -32.99         \\
CBST*~\cite{zou2018unsupervised}           & -0.36            & +0.58               & -1.00          & -1.12             & +0.88            & +1.69            & +1.03               & +0.28          \\

TENT~\cite{wang2020tent}          & +5.40           & -0.30              & -2.40         & -3.95              & -0.95             & +5.73           & +3.42              & +0.99  \\
ConjugatePL~\cite{goyal2022test}          & +5.93           & -0.03              &  -1.69        & -1.86              & +1.43             & +1.62           & +4.98             & +1.48 \\

GIPSO~\cite{saltori2022gipso}          & +13.12  & -0.54               & \textbf{+1.19} & \textbf{+2.45}    & +2.78            & +5.64            & +5.54               & +4.31          \\
\midrule
HGL(Ours)     & \textbf{+13.24}           & \textbf{+3.84}      & +0.79          & +1.95             & \textbf{+5.27}   & \textbf{+10.98}  & \textbf{+8.73}      & \textbf{+6.40} \\
\bottomrule
\end{tabular}
}
\label{table:syn4d2kitti}
\end{table*}

\noindent \textbf{Evaluation protocol:} For a fair comparison, we utilize the same evaluation metric as previous work~\cite{saltori2022gipso}. Specifically, we evaluate the model performance on a new incoming frame using the model adapted to the previous frame. We report the intersection-over-union (IoU) and the mean intersection-over-union (mIoU) improvement over the source model for effectiveness comparison.

\noindent \textbf{Implementation details:} We adopt the same network architecture as existing baseline methods. 
Specifically, we adopt the \textbf{MinkowskiNet}~\cite{choy20194d} as the backbone. For preparing the source model, here, we utilize the same training recipe as {GIPSO}~\cite{saltori2022gipso}.  
During online adaptation, the batch size is set to 1 for all benchmark datasets as current input. We apply the Adam optimizer with weight decay 1e-5 and set the learning rate to 1e-3 for all datasets. We abstain from employing schedulers because they necessitate prior information about the length of the data stream. For hyper-parameter, we set $\lambda=70, \alpha=0.99$ for all datasets.  For local K-NN, K is set to 10 for SemanticKITTI and 5 for nuScenes. As for $w$, we set it to 5 for SemanticKITTI and 1 for nuScenes following~\cite{saltori2022gipso}.

\begin{table*}[tbp]
\centering
\caption{mIoU improvement (\%) comparison in Synth4D to nuScenes test-time adaptation. $*$ denotes being re-purposed for test-time adaptation.}
\addtolength{\tabcolsep}{+2.0pt}
\resizebox{0.99\textwidth}{!}{
\begin{tabular}{lccccccca}
\toprule
\textbf{Model} & \textbf{Vehicle} & \textbf{Pedestrian} & \textbf{Road} & \textbf{Sidewalk} & \textbf{Terrain} & \textbf{Manmade} & \textbf{Vegetation} & \textbf{Avg} \\
\midrule
Source         & 22.45            & 14.38               & 42.03          & 28.39             & 15.58            & 38.18            & 54.14               & 30.75          \\
Upper Bound         & +3.76            & +0.92               & +9.41          & +16.95            & +19.79           & +10.92           & +10.71              & +10.35         \\
\midrule
ADABN~\cite{li2016revisiting}          & +1.23            & -2.74               & -1.24          & +0.14             & +0.53            & +0.70            & +4.03               & +0.38          \\
ProDA*~\cite{zhang2021prototypical}          & +0.57            & -1.40      & +0.73          & +0.09             & +0.71            & +0.40            & +0.91               & +0.29          \\
SHOT*~\cite{liang2020we}           & +0.82            & -1.77               & +0.68          & -0.05             & -0.70            & -0.54            & +1.09               & -0.07          \\
CBST*~\cite{zou2018unsupervised}           & +0.37            & -2.61               & -1.35          & -0.79             & +0.19            & -0.36            & -0.45               & -0.71          \\
TENT~\cite{wang2020tent}          & -0.16          & -0.20              & -1.25          & -0.29              & +0.02             & -0.12            & -0.34              & -0.34   \\
ConjugatePL~\cite{goyal2022test}          & +1.14           &\textbf{-0.41}              &+0.15         &+0.67              &+0.41             &+0.58            & +1.40             & +0.57    \\

GIPSO~\cite{saltori2022gipso}          & +0.55            & -3.76               & +1.64          & +1.72             & \textbf{+2.28}   & \textbf{+1.18}   & \textbf{+2.36}      & +0.85          \\
\midrule
HGL(Ours)     & \textbf{+1.42}   & -2.58               & \textbf{+5.57} & \textbf{+2.80}    & +2.16            & +1.02            & +2.32               & \textbf{+1.87} \\
\bottomrule
\end{tabular}   
}
\label{table:syn4d2nusc}
\end{table*}

\subsection{Experiment Results}
To verify the effectiveness of our HGL, we conduct extensive experiments across various adaptation scenarios, i.e., synthetic-to-real, real-to-real, different sensor configurations, continual learning, and adverse weather. Moreover, we further conduct experiments regarding efficiency.

\noindent \textbf{Results on synthetic-to-real setting:} We first conduct experiments in the synthetic-to-real setting, where synthetic data and real data share the same number of LiDAR beams. Results on SynLiDAR to SemanticKITTI, Synth4D to SemanticKITTI, and Synth4D to nuScenes are summarized in Table~\ref{table:syn2kitti}, \ref{table:syn4d2kitti} and \ref{table:syn4d2nusc}, respectively. The results in the table represent IoU improvements compared to the source model. The upper bound in the table means using ground truth as supervision.
As shown in Table~\ref{table:syn2kitti}, ~\ref{table:syn4d2kitti} and \ref{table:syn4d2nusc} our HGL achieves new state-of-the-art. Specifically, HGL obtains a mIoU improvement of +6.72\% on SynLiDAR to SemanticKITTI, +6.40\% on  Synth4D to SemanticKITTI and +1.87\% on Synth4D to nuScenes, with an improvement of 3.0\%, 2.1\% and 1.0\% compared to GIPSO. As shown in Table~\ref{table:syn2kitti} and Table~\ref{table:syn4d2kitti}, HGL also mitigates the negative transfer on imbalanced classes, i.e., the pedestrian class. 

\noindent {\textbf{Results on real-to-real setting:} In addition to the synthetic-to-real scenario, we also verify the effectiveness in real-to-real setting. The results on SemanticKITTI to nuScenes and nuScenes to SemanticKITTI are summarized in Table~{\ref{table:df_setting}}. An observation is that HGL consistently performs well in this setting.}

\noindent \textbf{Results on different sensor configurations:} We then conduct experiments on different sensor configurations, e.g., 32-beam to 64-beam and 64-beam to 32-beam. The Table~\ref{table:df_sensor} reports the results of Synth4D (32-beam) to SemanticKITTI, Synth4D (64-beam) to nuScenes, and SynLiDAR to nuScenes. As shown in Table~\ref{table:df_sensor}, HGL still achieves state-of-the-art. Specifically, HGL obtains a mIoU improvement of +8.7\% on  Synth4D (32-beam) to SemanticKITTI, with an improvement of 2.9\% compared to GIPSO.

\noindent \textbf{Results on continual learning:} We additionally conduct experiments on continual learning. The 150 sequences in nuScenes are derived from Singapore-Queenstown, Boston-Seaport, Singapore-Holland Village and Singapore-One, with variations among different sequences. Unlike the individual adaptations in normal, here we consider it as a continuous input stream for experiments in continual learning. Table~\ref{table:df_setting} demonstrates that despite the varying continuous inputs, HGL continues to exhibit significant improvements compared to GIPSO.

\noindent \textbf{Results on adverse weather:} We further verify the effectiveness of HGL on adverse weather. We utilized prior research~\cite{hahner2021fog, hahner2022lidar,dong2023benchmarking} to simulate adverse weather conditions such as fog and snow on the KITTI dataset by dividing the original sequence into $normal:fog:snow=1:2:2$. The results in Table ~\ref{table:df_setting} indicate that HGL can also excel in addressing extreme environments.

\begin{table}[t]
\centering
\addtolength{\tabcolsep}{+3.0pt}
\caption{mIoU (\%) results on real-to-real, continual learning and adverse weathers. The continual learning is on Synth4D to nuScenes, while the adverse weather is on SynLiDAR to SemanticKITTI. The adaptation time is calculated on SemanticKITTI.}
\resizebox{0.99\textwidth}{!}{
\begin{tabular}{lccccc}
\toprule
Setting                       & KITTI$\rightarrow$nuScenes &nuScenes$\rightarrow$KITTI    & Continual learning & Adverse weather  & Time(s) \\
\midrule
Source     & 32.03 & 37.40 & 30.75  & 36.66  & -      \\
GIPSO      &33.35(+1.05) & 40.28(+2.88) & 31.17(+0.42)  & 39.71(+3.05)  & 3.83     \\
\rowcolor{Gray} HGL(Ours)  & 34.49(+2.46) & 45.06(+7.66)  & 32.07(+1.32)  & 42.57(+5.91)  & 0.63     \\
\bottomrule
\end{tabular}
}
\label{table:df_setting}
\end{table}

\subsection{Experimental Analysis}

\par  We conduct an extensive experimental analysis in Table~\ref{table:ablation} and Table~\ref{table:novelty} to show the contribution of each component. Unless stated otherwise, the experiments are based on SynLiDAR to SemanticKITTI. 

\begin{figure}[t]
\centering
\begin{minipage}[t]{0.48\textwidth}
\centering
\includegraphics[width=0.99\textwidth]{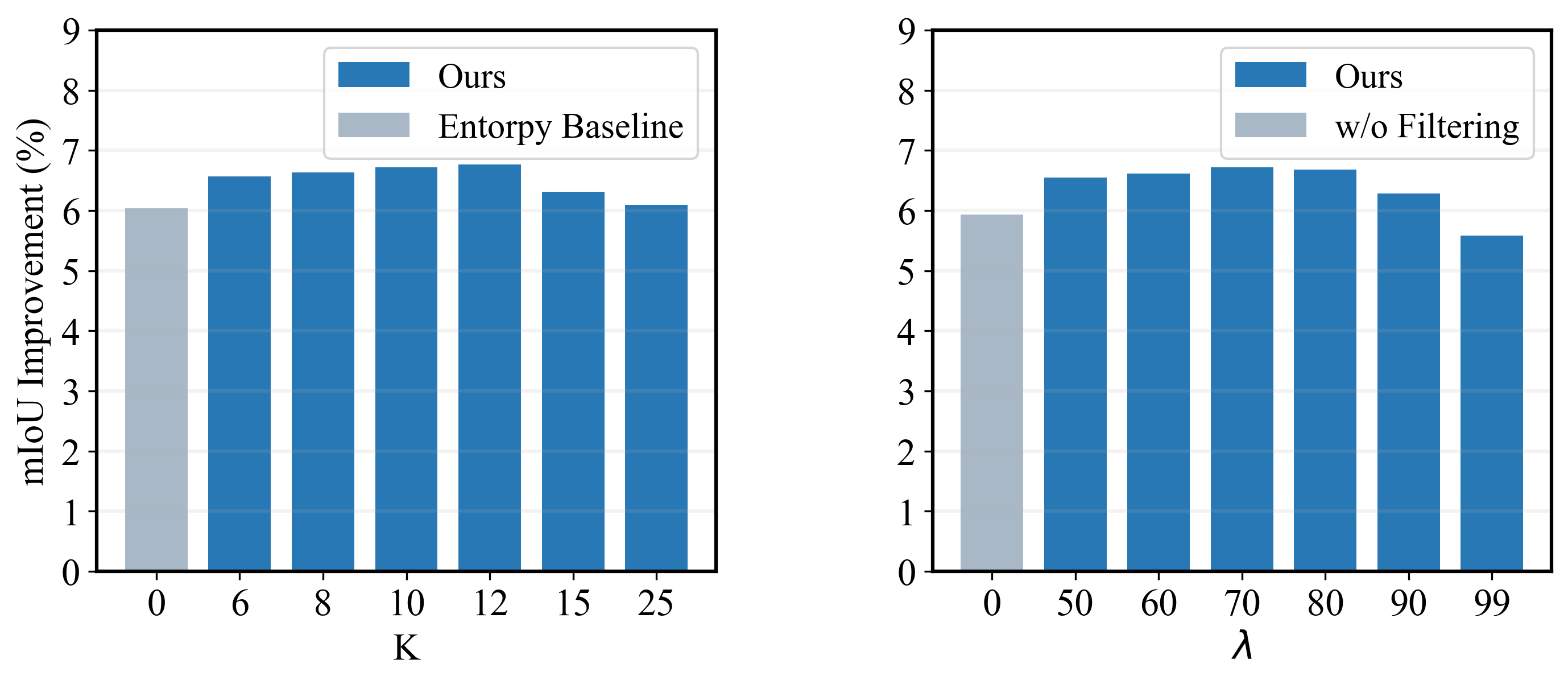}
\caption{Hyper-parameter sensitivity analysis on SynLiDAR to SemanticKITTI.}
\label{fig:hyper_param}
\end{minipage}
\begin{minipage}[t]{0.48\textwidth}
\centering
\includegraphics[width=0.99\textwidth]{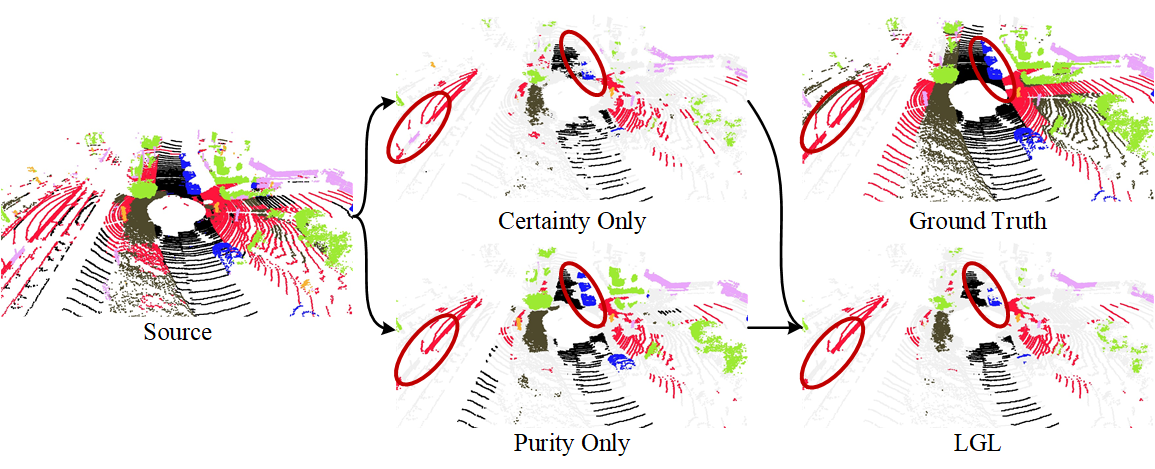}
\caption{Visualization of selected pseudo-label of LGL.}
\label{fig:LGL}
\end{minipage}
\end{figure}

\noindent \textbf{Hyper-parameter Sensitivity:} We first study the parameter sensitivity of $K$ in Fig.~\ref{fig:hyper_param} left, where $K$ is in the range of [0, 25]. Note that $K = 0$ denotes that we do not introduce the local geometry learning. It is easy to find that results around the selected parameter $K = 10$  are stable, and much better than the source model. In Fig.~\ref{fig:hyper_param} right, we present the mIoU for $\lambda$. $\lambda=0$ indicates no filtering of local geometric pseudo labels. For all benchmarks, we pre-define $\lambda = 70$ to separate ``reliable" and ``unreliable" local geometric pseudo-label. With smaller step sizes, we may be able to find better hyperparameters. The results show that mIoU is relatively stable around our selection.

\noindent \textbf{Local Geometry Learning (LGL):} To assess the effectiveness of the proposed Local Geometry Learning (LGL), we remove the neighboring aggregation and solely employ the entropy of single-point as the confidence to select pseudo-labels. In Table~\ref{table:ablation}, experiment VI shows that the model's performance improvement is limited. 
Furthermore, we present the visualization of the pseudo-label of each component of LGL. As shown in Fig.\ref{fig:LGL}, although the neighboring aggregation and prediction certainty achieve impressive results with the introduction of local geometric information, it still has problems in some details. Geometric purity complements prediction certainty by enhancing the spatial proximity property, resulting in robust pseudo-labels.

\begin{table}[tp]
\begin{minipage}{.47\linewidth}
  \centering
  \caption{mIoU (\%) comparison of methods for different sensor configurations.}
  \addtolength{\tabcolsep}{-0.0pt}
  \resizebox{1.0\textwidth}{!}{
    \begin{tabular}{c|c|c|c}
        \toprule
        Source                       & Target                                         & Model                           & mIoU  \\ \hline
        \multirow{4}{*}{Synth4D(32)} & \multicolumn{1}{c|}{nuScenes(32)}                  & \multicolumn{1}{c|}{Source}     & 30.75 \\ \cline{2-4} 
                                     & \multicolumn{1}{c|}{\multirow{3}{*}{SemanticKITTI(64)}}    & \multicolumn{1}{c|}{Source}     & 27.55 \\
                                     & \multicolumn{1}{c|}{}                          & \multicolumn{1}{c|}{GIPSO}      & 33.36(+5.81) \\
                                     & \multicolumn{1}{c|}{}                          & \multicolumn{1}{c|}{HGL} & 36.25(+8.70) \\ \hline
        \multirow{4}{*}{Synth4D(64)} & \multicolumn{1}{c|}{SemanticKITTI(64)}                     & \multicolumn{1}{c|}{Source}     & 35.93 \\ \cline{2-4} 
                                     & \multicolumn{1}{c|}{\multirow{3}{*}{nuScenes(32)}} & \multicolumn{1}{c|}{Source}     & 30.14 \\
                                     & \multicolumn{1}{c|}{}                          & \multicolumn{1}{c|}{GIPSO}      & 28.88(-1.26) \\
                                     & \multicolumn{1}{c|}{}                          & \multicolumn{1}{c|}{HGL} & 29.77(-0.37) \\ \hline
        \multirow{4}{*}{SynLiDAR(64)}    & \multicolumn{1}{c|}{SemanticKITTI(64)}                     & \multicolumn{1}{c|}{Source}     & 40.19 \\ \cline{2-4} 
                                     & \multicolumn{1}{c|}{\multirow{3}{*}{nuScenes(32)}} & \multicolumn{1}{c|}{Source}     & 32.45 \\
                                     & \multicolumn{1}{c|}{}                          & \multicolumn{1}{c|}{GIPSO}      & 32.52(+0.07) \\
                                     & \multicolumn{1}{c|}{}                          & \multicolumn{1}{c|}{HGL} & 32.85(+0.40) \\ \bottomrule
        \end{tabular}
    }
  \label{table:df_sensor}%
\end{minipage}
\ \ \ 
\begin{minipage}{.47\linewidth}
  \centering
  \caption{Ablation study in SynLiDAR to SemanticKITTI test-time adaptation.}
  \resizebox{1.0\textwidth}{!}{
    \begin{tabular}{c|ccccc|cc}
    \toprule
    ID     & LGL        & TGR        & GGF        & CW        & ALG        & mIoU(\%)  &Time(s)\\
    \midrule
    Source &   &            &            &            &            & 40.19 &-\\ 
    \midrule
    I & \checkmark &            &            &            &            & +0.63 &0.518\\
    II      & \checkmark & \checkmark &            &            &            & +3.30 &0.612\\
    III     & \checkmark & \checkmark & \checkmark &            &            & +5.40 &0.628\\
    IV    & \checkmark & \checkmark & \checkmark & \checkmark &            & +6.12 &0.628\\
    V     & \checkmark & \checkmark & \checkmark & \checkmark & \checkmark & +6.72 &0.630\\
    \midrule
    VI     &    & \checkmark & \checkmark & \checkmark & \checkmark & +6.04 &0.606\\
    VII     & \checkmark &  & \checkmark & \checkmark & \checkmark & +2.01 &0.535\\
    VIII     & \checkmark & \checkmark &  & \checkmark &\checkmark  & +2.20 &0.612\\
    IX     & \checkmark & \checkmark & \checkmark &  & \checkmark & +5.10 &0.631\\
    \bottomrule
    \end{tabular}
    }
  \label{table:ablation}%
\end{minipage}
\end{table}%

\begin{table}[t]
\centering
\addtolength{\tabcolsep}{+3.0pt}
\caption{{mIoU (\%) improvement of different implementations within each module compared to GIPSO.}}
\resizebox{0.99\textwidth}{!}{
\begin{tabular}{l|ccc|ccc|cc}
\toprule
           & \multicolumn{3}{c}{Global Geometry Fine-tuning}                & \multicolumn{3}{|c|}{Temporal Regularization} & \multicolumn{2}{c}{GIPSO} \\
           & ProDA & Weighted-Clustering & BMD & SimSiam         &CosSim & KL      & Vanilla  & +HGL                         \\
\hline
SynL$\rightarrow$KITTI & +3.02   & +4.23          & +2.77              & +3.02         & +0.85     & +1.52      & +0.00     & +2.15                       \\
Synt$\rightarrow$KITTI & +2.09   & +2.06          & +3.00              & +2.09         & +0.82     & +0.98      & +0.00     & +1.79                           \\
Synt$\rightarrow$nuSc  & +1.02   & +0.91          & +0.96             & +1.02         & +0.21     & +0.19      & +0.00     & +0.38              \\
\bottomrule
\end{tabular}
}
\label{table:novelty}
\end{table}

\noindent \textbf{Global Geometry Fine-tuning (GGF):} We ablate Global Geometry Fine-tuning (GGF) in Table~\ref{table:ablation} experiment II and III, V and VIII. Results show that the addition of GGF resulted in a slight increase in run time, but a significant performance improvement. V and VIII show that GGF successfully introduces a progressively increasing target information and further improves performance by 4.5 points. Note that the low performance in VIII is due to the indiscriminate use of all local geometric pseudo-labels as supervisory. 

\noindent \textbf{Temporal Geometry Regularization (TGR):} We show the effect of Temporal Geometry Regularization (TGR) in Table~\ref{table:ablation}. The comparisons between experiments I and II, V and VII demonstrate that the proposed TGR is very crucial due to the sensitivity of the TTA, which makes the whole optimization process stable and efficient.

\noindent \textbf{Confidence Weight (CW):} In Table~\ref{table:ablation}, we compare the performance with and without the confidence weight in experiments V and IX, which show the contribution of the confidence weight in reducing negative transfer.

\noindent \textbf{Fine-tuning on All Local Geometric Pseudo-label (ALG):} The fine-tuning on all local geometric pseudo-labels achieved a balance between the precision and generalization of the pseudo-labels. To illustrate this, in Table~\ref{table:ablation}, we compare the model with only fine-tuning on selected geometric pseudo-labels(IV and V). It shows that simply using the latter worsens the performance from +6.72\% to +6.12\%.

\noindent \textcolor{black}{\textbf{Flexibility:}  It is noteworthy that the specific implementations within each module are not firmly fixed and can be updated to incorporate advancements in the field. {For instance, at the global geometry learning level, as demonstrated in Table~{\ref{table:novelty}}, HGL extends beyond prototype modeling like ProDA~\cite{zhang2021prototypical}. We can also explore constructing prototypes using weighted clustering or BMD~\cite{qu2022bmd}. At the temporal level, alternatives to SimSiam~\cite{chen2021exploring}, simple cosine similarity, and the KL divergence, prove to be not only feasible but also superior to GIPSO.} Moreover, integrating HGL into GIPSO to build a thorough local-global-temporal structure results in notable improvements. In summary, Table~{\ref{table:novelty}} supports the flexibility and generalization of our HGL framework. 
}

\begin{figure}[t]
    \centering
    \includegraphics[width=0.95\textwidth]{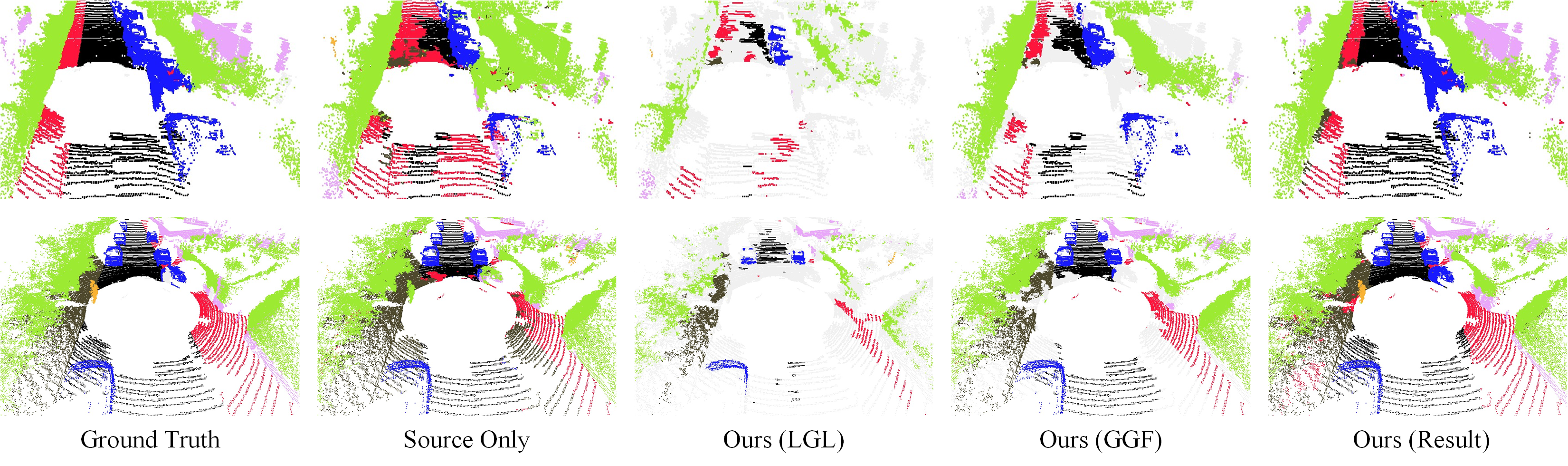}
    \caption{Visualization of selected pseudo-label and adaptation results for SynLiDAR to SemanticKITTI. From left to right: ground-truth label, pseudo-label generated by Source Only model, selected local pseudo-label generated by Our LGL, local-global pseudo-label fine-tuned by Our GGF, and adaptation result predicted by Ours HGL.}
    \label{fig:compare_our}
\end{figure}

\begin{figure}[t]
    \centering
\    \includegraphics[width=0.95\textwidth]{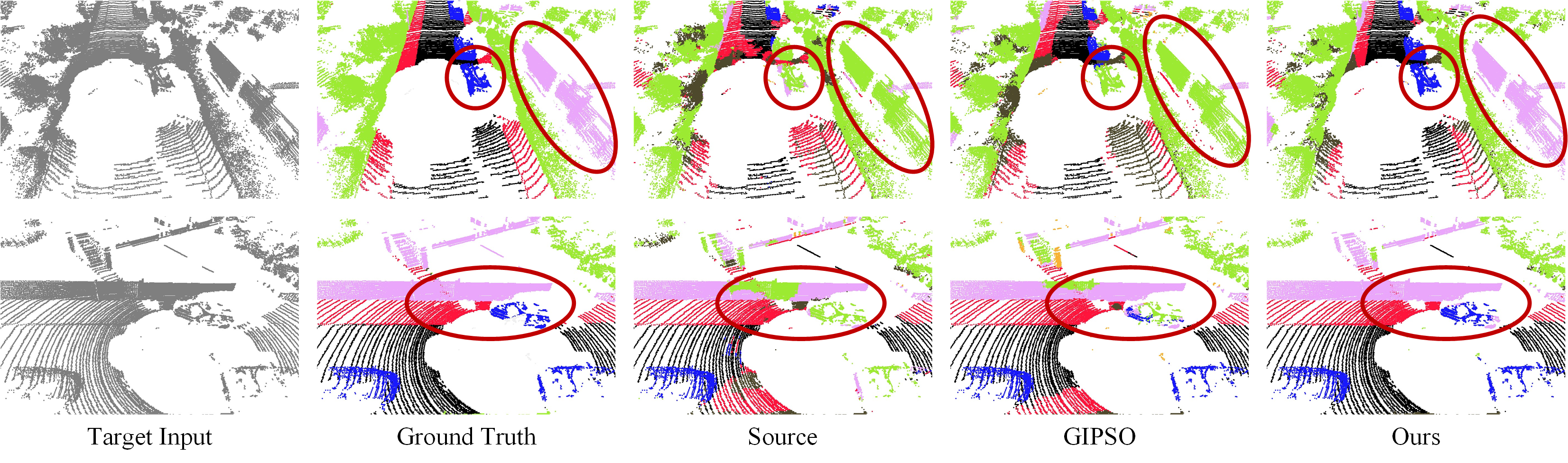}
    \caption{{Visual comparison between GIPSO and our HGL. From left to right: target input, ground-truth label, result predicted by source model, adaptation result predicted by GIPSO and our HGL.}}
    \label{fig:compare_gipso}
\end{figure}

\noindent \textbf{Efficiency:} The last column of Table~\ref{table:df_setting} exhibits the adaptation time. Compared to GPISO, our approach reduces the adaptation time by 80\%, which is mainly attributed to that our HGL does not introduce additional networks to extract geometric information and does not perform multiple calculations to estimate the uncertainty of pseudo-labels.

\noindent \textbf{Qualitative results:} Fig.~\ref{fig:compare_our} depicts the pseudo-label and segmentation result of HGL. It is clear to find that LGL can accurately select pseudo-labels, while GGF can generalize them to a wider range of points. Fig.~\ref{fig:compare_gipso} shows the comparison between HGL, GIPSO, and the source model on SynLiDAR to SemanticKITTI. We observe that the results of our GLC are generally superior to GIPSO.

\section{Conclusion}
\par In this paper, we have presented \emph{Hierarchical Geometry Learning (HGL) framework} for test-time adaptation in 3D LiDAR segmentation. 
Technically, we have devised an innovative local geometry learning module to generate pseudo-label, proposed a global geometry fine-tuning module for pseudo-label fine-tuning, and introduced a temporal geometry regularization module for consistency regularization.
Compared to existing approaches lacking thorough exploration of geometry information, HGL is appealing by the geometry hierarchical structure from local (point-level), global (object-level), and temporal (frame-level) for accurate and robust self-supervised learning in TTA-3DSeg.
Extensive experiments verify the high adaptability and efficiency of HGL.  Remarkably, on the SynLiDAR to SemanticKITTI task,  HGL significantly outperforms GIPSO by 3.0\% and significantly reduces the required adaptation time by 80\%. 

\par \noindent \textbf{Acknowledgments}: This work is supported by the National Natural Science Foundation of China (No. 62372329), in part by Shanghai Scientific Innovation Foundation (No.23DZ1203400), in part by Shanghai Rising Star Program (No.21QC1400900), in part by Tongji-Qomolo Autonomous Driving Commercial Vehicle Joint Lab Project, and in part by Xiaomi Young Talents Program.

%
%
\bibliographystyle{splncs04}
\bibliography{main}

\begin{thebibliography}{10}
\providecommand{\url}[1]{\texttt{#1}}
\providecommand{\urlprefix}{URL }
\providecommand{\doi}[1]{https://doi.org/#1}

\bibitem{balsiger2019learning}
Balsiger, F., Soom, Y., Scheidegger, O., Reyes, M.: Learning shape representation on sparse point clouds for volumetric image segmentation. In: Medical Image Computing and Computer Assisted Intervention--MICCAI 2019: 22nd International Conference, Shenzhen, China, October 13--17, 2019, Proceedings, Part II 22. pp. 273--281. Springer (2019)

\bibitem{behley2019semantickitti}
Behley, J., Garbade, M., Milioto, A., Quenzel, J., Behnke, S., Stachniss, C., Gall, J.: Semantickitti: A dataset for semantic scene understanding of lidar sequences. In: Proceedings of the IEEE/CVF international conference on computer vision. pp. 9297--9307 (2019)

\bibitem{caesar2020nuscenes}
Caesar, H., Bankiti, V., Lang, A.H., Vora, S., Liong, V.E., Xu, Q., Krishnan, A., Pan, Y., Baldan, G., Beijbom, O.: nuscenes: A multimodal dataset for autonomous driving. In: Proceedings of the IEEE/CVF conference on computer vision and pattern recognition. pp. 11621--11631 (2020)

\bibitem{chen2022contrastive}
Chen, D., Wang, D., Darrell, T., Ebrahimi, S.: Contrastive test-time adaptation. In: Proceedings of the IEEE/CVF Conference on Computer Vision and Pattern Recognition. pp. 295--305 (2022)

\bibitem{chen2021exploring}
Chen, X., He, K.: Exploring simple siamese representation learning. In: Proceedings of the IEEE/CVF conference on computer vision and pattern recognition. pp. 15750--15758 (2021)

\bibitem{choy20194d}
Choy, C., Gwak, J., Savarese, S.: 4d spatio-temporal convnets: Minkowski convolutional neural networks. In: Proceedings of the IEEE/CVF conference on computer vision and pattern recognition. pp. 3075--3084 (2019)

\bibitem{dong2023benchmarking}
Dong, Y., Kang, C., Zhang, J., Zhu, Z., Wang, Y., Yang, X., Su, H., Wei, X., Zhu, J.: Benchmarking robustness of 3d object detection to common corruptions in autonomous driving. arXiv preprint arXiv:2303.11040  (2023)

\bibitem{dosovitskiy2020image}
Dosovitskiy, A., Beyer, L., Kolesnikov, A., Weissenborn, D., Zhai, X., Unterthiner, T., Dehghani, M., Minderer, M., Heigold, G., Gelly, S., et~al.: An image is worth 16x16 words: Transformers for image recognition at scale. arXiv preprint arXiv:2010.11929  (2020)

\bibitem{dosovitskiy2017carla}
Dosovitskiy, A., Ros, G., Codevilla, F., Lopez, A., Koltun, V.: Carla: An open urban driving simulator. In: Conference on robot learning. pp. 1--16. PMLR (2017)

\bibitem{goyal2022test}
Goyal, S., Sun, M., Raghunathan, A., Kolter, J.Z.: Test time adaptation via conjugate pseudo-labels. Advances in Neural Information Processing Systems  \textbf{35},  6204--6218 (2022)

\bibitem{hahner2022lidar}
Hahner, M., Sakaridis, C., Bijelic, M., Heide, F., Yu, F., Dai, D., Van~Gool, L.: Lidar snowfall simulation for robust 3d object detection. In: Proceedings of the IEEE/CVF Conference on Computer Vision and Pattern Recognition. pp. 16364--16374 (2022)

\bibitem{hahner2021fog}
Hahner, M., Sakaridis, C., Dai, D., Van~Gool, L.: Fog simulation on real lidar point clouds for 3d object detection in adverse weather. In: Proceedings of the IEEE/CVF International Conference on Computer Vision. pp. 15283--15292 (2021)

\bibitem{hatem2023point}
Hatem, A., Qian, Y., Wang, Y.: Point-tta: Test-time adaptation for point cloud registration using multitask meta-auxiliary learning. In: Proceedings of the IEEE/CVF International Conference on Computer Vision. pp. 16494--16504 (2023)

\bibitem{hatem2023test}
Hatem, A., Qian, Y., Wang, Y.: Test-time adaptation for point cloud upsampling using meta-learning. In: 2023 IEEE/RSJ International Conference on Intelligent Robots and Systems (IROS). pp. 1284--1291. IEEE (2023)

\bibitem{he2016deep}
He, K., Zhang, X., Ren, S., Sun, J.: Deep residual learning for image recognition. In: Proceedings of the IEEE conference on computer vision and pattern recognition. pp. 770--778 (2016)

\bibitem{he2023density}
He, Y., Jin, L., Guo, B., Huo, Z., Wang, H., Jin, Q.: Density-based road segmentation algorithm for point cloud collected by roadside lidar. Automotive Innovation  \textbf{6}(1),  116--130 (2023)

\bibitem{jadon2020survey}
Jadon, S.: A survey of loss functions for semantic segmentation. In: 2020 IEEE conference on computational intelligence in bioinformatics and computational biology (CIBCB). pp.~1--7. IEEE (2020)

\bibitem{kim2023single}
Kim, H., Kang, Y., Oh, C., Yoon, K.J.: Single domain generalization for lidar semantic segmentation. In: Proceedings of the IEEE/CVF Conference on Computer Vision and Pattern Recognition. pp. 17587--17598 (2023)

\bibitem{kim2022ev}
Kim, J., Hwang, I., Kim, Y.M.: Ev-tta: Test-time adaptation for event-based object recognition. In: Proceedings of the IEEE/CVF Conference on Computer Vision and Pattern Recognition. pp. 17745--17754 (2022)

\bibitem{langer2020domain}
Langer, F., Milioto, A., Haag, A., Behley, J., Stachniss, C.: Domain transfer for semantic segmentation of lidar data using deep neural networks. In: 2020 IEEE/RSJ International Conference on Intelligent Robots and Systems (IROS). pp. 8263--8270. IEEE (2020)

\bibitem{li2016revisiting}
Li, Y., Wang, N., Shi, J., Liu, J., Hou, X.: Revisiting batch normalization for practical domain adaptation. arXiv preprint arXiv:1603.04779  (2016)

\bibitem{liang2020we}
Liang, J., Hu, D., Feng, J.: Do we really need to access the source data? source hypothesis transfer for unsupervised domain adaptation. In: International conference on machine learning. pp. 6028--6039. PMLR (2020)

\bibitem{liu2023grab}
Liu, Y., Li, W., Liu, J., Chen, H., Yuan, Y.: Grab-net: Graph-based boundary-aware network for medical point cloud segmentation. IEEE Transactions on Medical Imaging  (2023)

\bibitem{mirza2022norm}
Mirza, M.J., Micorek, J., Possegger, H., Bischof, H.: The norm must go on: Dynamic unsupervised domain adaptation by normalization. In: Proceedings of the IEEE/CVF Conference on Computer Vision and Pattern Recognition. pp. 14765--14775 (2022)

\bibitem{mugnai2021laser}
Mugnai, F.: Laser scanning and point cloud segmentation for contactless geo-mechanical surveying: Conservative restoration in hypogeum environment. The International Archives of the Photogrammetry, Remote Sensing and Spatial Information Sciences  \textbf{46},  455--461 (2021)

\bibitem{niu2022efficient}
Niu, S., Wu, J., Zhang, Y., Chen, Y., Zheng, S., Zhao, P., Tan, M.: Efficient test-time model adaptation without forgetting. In: International conference on machine learning. pp. 16888--16905. PMLR (2022)

\bibitem{qi2017pointnet}
Qi, C.R., Su, H., Mo, K., Guibas, L.J.: Pointnet: Deep learning on point sets for 3d classification and segmentation. In: Proceedings of the IEEE conference on computer vision and pattern recognition. pp. 652--660 (2017)

\bibitem{qin2019pointdan}
Qin, C., You, H., Wang, L., Kuo, C.C.J., Fu, Y.: Pointdan: A multi-scale 3d domain adaption network for point cloud representation. Advances in Neural Information Processing Systems  \textbf{32} (2019)

\bibitem{qu2022bmd}
Qu, S., Chen, G., Zhang, J., Li, Z., He, W., Tao, D.: Bmd: A general class-balanced multicentric dynamic prototype strategy for source-free domain adaptation. In: European Conference on Computer Vision. pp. 165--182. Springer (2022)

\bibitem{qu2023modality}
Qu, S., Pan, Y., Chen, G., Yao, T., Jiang, C., Mei, T.: Modality-agnostic debiasing for single domain generalization. In: Proceedings of the IEEE/CVF Conference on Computer Vision and Pattern Recognition. pp. 24142--24151 (2023)

\bibitem{qu2024lead}
Qu, S., Zou, T., He, L., R{\"o}hrbein, F., Knoll, A., Chen, G., Jiang, C.: Lead: Learning decomposition for source-free universal domain adaptation. In: Proceedings of the IEEE/CVF Conference on Computer Vision and Pattern Recognition. pp. 23334--23343 (2024)

\bibitem{qu2023upcycling}
Qu, S., Zou, T., R{\"o}hrbein, F., Lu, C., Chen, G., Tao, D., Jiang, C.: Upcycling models under domain and category shift. In: Proceedings of the IEEE/CVF Conference on Computer Vision and Pattern Recognition. pp. 20019--20028 (2023)

\bibitem{saltori2022gipso}
Saltori, C., Krivosheev, E., Lathuili{\'e}re, S., Sebe, N., Galasso, F., Fiameni, G., Ricci, E., Poiesi, F.: Gipso: Geometrically informed propagation for online adaptation in 3d lidar segmentation. In: European Conference on Computer Vision. pp. 567--585. Springer (2022)

\bibitem{saltori2023walking}
Saltori, C., Osep, A., Ricci, E., Leal-Taix{\'e}, L.: Walking your lidog: A journey through multiple domains for lidar semantic segmentation. In: Proceedings of the IEEE/CVF International Conference on Computer Vision. pp. 196--206 (2023)

\bibitem{sanchez2023domain}
Sanchez, J., Deschaud, J.E., Goulette, F.: Domain generalization of 3d semantic segmentation in autonomous driving. In: Proceedings of the IEEE/CVF International Conference on Computer Vision. pp. 18077--18087 (2023)

\bibitem{shaban2023lidar}
Shaban, A., Lee, J., Jung, S., Meng, X., Boots, B.: Lidar-uda: Self-ensembling through time for unsupervised lidar domain adaptation. In: Proceedings of the IEEE/CVF International Conference on Computer Vision. pp. 19784--19794 (2023)

\bibitem{shannon1948mathematical}
Shannon, C.E.: A mathematical theory of communication. The Bell system technical journal  \textbf{27}(3),  379--423 (1948)

\bibitem{shin2022mm}
Shin, I., Tsai, Y.H., Zhuang, B., Schulter, S., Liu, B., Garg, S., Kweon, I.S., Yoon, K.J.: Mm-tta: multi-modal test-time adaptation for 3d semantic segmentation. In: Proceedings of the IEEE/CVF Conference on Computer Vision and Pattern Recognition. pp. 16928--16937 (2022)

\bibitem{shu2023clipood}
Shu, Y., Guo, X., Wu, J., Wang, X., Wang, J., Long, M.: Clipood: Generalizing clip to out-of-distributions. arXiv preprint arXiv:2302.00864  (2023)

\bibitem{veksler2023test}
Veksler, O.: Test time adaptation with regularized loss for weakly supervised salient object detection. In: Proceedings of the IEEE/CVF Conference on Computer Vision and Pattern Recognition. pp. 7360--7369 (2023)

\bibitem{vs2023towards}
VS, V., Oza, P., Patel, V.M.: Towards online domain adaptive object detection. In: Proceedings of the IEEE/CVF Winter Conference on Applications of Computer Vision. pp. 478--488 (2023)

\bibitem{wang2023space}
Wang, C., Xie, H., Yuan, Y., Fu, C., Yue, X.: Space engage: Collaborative space supervision for contrastive-based semi-supervised semantic segmentation. In: Proceedings of the IEEE/CVF International Conference on Computer Vision. pp. 931--942 (2023)

\bibitem{wang2020tent}
Wang, D., Shelhamer, E., Liu, S., Olshausen, B., Darrell, T.: Tent: Fully test-time adaptation by entropy minimization. arXiv preprint arXiv:2006.10726  (2020)

\bibitem{wang2023dynamically}
Wang, W., Zhong, Z., Wang, W., Chen, X., Ling, C., Wang, B., Sebe, N.: Dynamically instance-guided adaptation: A backward-free approach for test-time domain adaptive semantic segmentation. In: Proceedings of the IEEE/CVF Conference on Computer Vision and Pattern Recognition. pp. 24090--24099 (2023)

\bibitem{wu2018squeezeseg}
Wu, B., Wan, A., Yue, X., Keutzer, K.: Squeezeseg: Convolutional neural nets with recurrent crf for real-time road-object segmentation from 3d lidar point cloud. In: 2018 IEEE international conference on robotics and automation (ICRA). pp. 1887--1893. IEEE (2018)

\bibitem{wu2019squeezesegv2}
Wu, B., Zhou, X., Zhao, S., Yue, X., Keutzer, K.: Squeezesegv2: Improved model structure and unsupervised domain adaptation for road-object segmentation from a lidar point cloud. In: 2019 international conference on robotics and automation (ICRA). pp. 4376--4382. IEEE (2019)

\bibitem{xiao2021synlidar}
Xiao, A., Huang, J., Guan, D., Zhan, F., Lu, S.: Synlidar: Learning from synthetic lidar sequential point cloud for semantic segmentation. arXiv preprint arXiv:2107.05399  \textbf{1} (2021)

\bibitem{yi2021complete}
Yi, L., Gong, B., Funkhouser, T.: Complete \& label: A domain adaptation approach to semantic segmentation of lidar point clouds. In: Proceedings of the IEEE/CVF conference on computer vision and pattern recognition. pp. 15363--15373 (2021)

\bibitem{zhang2021prototypical}
Zhang, P., Zhang, B., Zhang, T., Chen, D., Wang, Y., Wen, F.: Prototypical pseudo label denoising and target structure learning for domain adaptive semantic segmentation. In: Proceedings of the IEEE/CVF conference on computer vision and pattern recognition. pp. 12414--12424 (2021)

\bibitem{zhang2022auxadapt}
Zhang, Y., Borse, S., Cai, H., Porikli, F.: Auxadapt: Stable and efficient test-time adaptation for temporally consistent video semantic segmentation. In: Proceedings of the IEEE/CVF Winter Conference on Applications of Computer Vision. pp. 2339--2348 (2022)

\bibitem{zhao2021epointda}
Zhao, S., Wang, Y., Li, B., Wu, B., Gao, Y., Xu, P., Darrell, T., Keutzer, K.: epointda: An end-to-end simulation-to-real domain adaptation framework for lidar point cloud segmentation. In: Proceedings of the AAAI Conference on Artificial Intelligence. vol.~35, pp. 3500--3509 (2021)

\bibitem{zou2018unsupervised}
Zou, Y., Yu, Z., Kumar, B., Wang, J.: Unsupervised domain adaptation for semantic segmentation via class-balanced self-training. In: Proceedings of the European conference on computer vision (ECCV). pp. 289--305 (2018)

\end{thebibliography}
\end{document}


\title{HGL: Hierarchical Geometry Learning for Test-time Adaptation in 3D Point Cloud Segmentation} 

\titlerunning{HGL: Test-time Adaptation in Point Cloud Segmentation}

\author{Tianpei Zou\inst{1}\thanks{Equal Contribution}\ \and Sanqing Qu \inst{1}\repeatthanks \and Zhijun Li \inst{1}\and \\
Alois Knoll \inst{2} \and Lianghua He \inst{1} \and Guang Chen\inst{1}$^{(\textrm{\Letter})}$ \and Changjun Jiang\inst{1} }

\authorrunning{T.~Zou and S.~Qu et al.}

\institute{Tongji University, \and Technical University of Munich \\ \email{\{2011459, 2011444, guangchen\}@tongji.edu.cn}} 

\clearpage
\setcounter{page}{1}
\maketitle


\section{More Details about Methodology}
\subsection{Temporal Geometry Regularization}
We start by calculating the geometry corresponding points between $X_T^{t-w}$ and $X_T^t$ using the rigid transformation $T^{t-w \rightarrow t}$ from the odometry. The geometry corresponding points $\Theta^{t-w \rightarrow t}$ are defined as:
\begin{equation}
\begin{aligned}
\Theta^{t, t-w}= & \left\{\left\{\mathrm{x}^t \in X_{\mathcal{T}}^t, \mathrm{x}^{t-w} \in X_{\mathcal{T}}^{t-w}\right\}:\right. \\
& \mathrm{x}^t=\mathrm{NN}\left(T^{t-w \rightarrow t}  \mathrm{x}^{t-w}, X_{\mathcal{T}}^t\right), \\
& \left.\left\|\mathrm{x}^t-\mathrm{x}^{t-w}\right\|_2<\tau\right\}
\end{aligned}
\end{equation}
where $\mathrm{NN}(n,m)$ is the nearest-neighbor of $n$ to $m$ and $\tau$ is a distance threshold. Following SimSiam~\cite{chen2021exploring}, an encoder network $h(.)$ and a predictor head $f(.)$ are introduced to the target model $F_T$ and aim at minimizing the negative cosine similarity of semantic representations of geometry corresponding points. Furthermore, we demand the alignment of low-confidence features with their high-confidence counterparts, which effectively prevents the corrupting of high-confidence features by low-confidence ones. The negative cosine similarity can be formulated as:
\begin{equation}
    \mathcal{D}^{\prime}_{t \rightarrow t-w}\left(q^t, z^{t-w}\right)=-S^{t-w}\frac{q^t}{\left\|q^t\right\|_2} \cdot \frac{z^{t-w}}{\left\|z^{t-w}\right\|_2} 
\end{equation}
where $z^t=h(x^t)$ is the encoder features and $q^t=f(h(x^t))$ is the respective predictor features. The final temporal consistency loss is:
\begin{equation}
    \mathcal{L}_{reg}=\frac{1}{2} \mathcal{D}^{\prime}_{t \rightarrow t-w}\left(q^t, z^{t-w}\right)+\frac{1}{2} \mathcal{D}^{\prime}_{t-w \rightarrow t}\left(q^{t-w}, z^t\right)
\end{equation}
where \texttt{stop-grad} operator is applied to $z^{t}$ and $z^{t-w}$.

\subsection{Soft Dice Loss}
As aforementioned, class imbalances are typically pronounced in point clouds. Furthermore, the global geometry fine-tuning may introduce some erroneous low-confidence pseudo-labels. To alleviate the impact of category imbalance and incorrect pseudo-label, we employ soft Dice loss $\mathcal{L}_{dice}$~\cite{jadon2020survey} and smooth labels~\cite{liang2020we}. Formally, the $\mathcal{L}_{dice}$ is defined as:
\begin{equation}
\mathcal{L}_{dice}(y, \hat{p})=1-\frac{2\cdot y\cdot \hat{p}}{y+\hat{p}}
\end{equation}
Here, $y$ is the predicted value by the target model, $\hat{p}$  is the smoothed one-hot encoding i.e., $\hat{p} = (1 -\beta)*\mathbf{1}_{[k=y_s]} + \beta / K$, and $\beta$ is the smoothing parameter. Different from~\cite{liang2020we} which uses a fixed $\beta$, we use confidence to adaptively adjust $\beta=\hat{\beta}(1-S)$, where $\hat{\beta}$ is empirically set to 0.3.


\begin{figure}[t]
    \centering
    \includegraphics[width=0.80\textwidth]{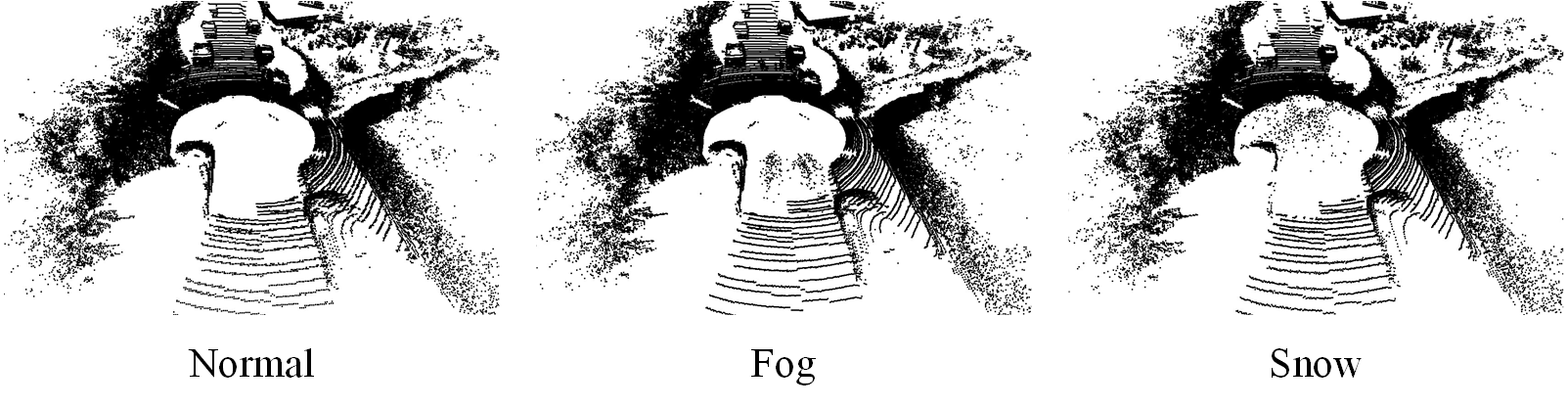}
    \caption{Visualization of simulated adverse weathers for SemanticKITTI. From left to right: the normal point cloud, the fog point cloud and the snow point cloud are shown one by one.}
    \label{fig:Adverse_Weathers}
\end{figure}

\section{More Experiment Details}
\subsection{Adverse Weathers Details}
\par We utilized prior research~\cite{hahner2021fog, hahner2022lidar,dong2023benchmarking} to simulate adverse weather such as fog and snow on the SemanticKITTI dataset. \cite{hahner2021fog} and \cite{hahner2022lidar} first proposed a physically based approach to point cloud fog and snow simulation. \cite{dong2023benchmarking} integrates the two approaches and provides five different levels to represent the severity of adverse weather. To maximize the exploration of how our method performs in adverse weather, we uniformly choose the highest level and divide the original sequence into $normal:fog:snow=1:2:2$.  For the semantic labels, we do not change the labels of slightly fluctuating points and set the labels of drastically changing or regenerating points to unlabeled. The visualization of point cloud fog and snow simulation is shown in Fig. \ref{fig:Adverse_Weathers}.

\begin{figure}[t]
\centering
\begin{minipage}[t]{0.48\textwidth}
\centering
\includegraphics[width=0.99\textwidth]{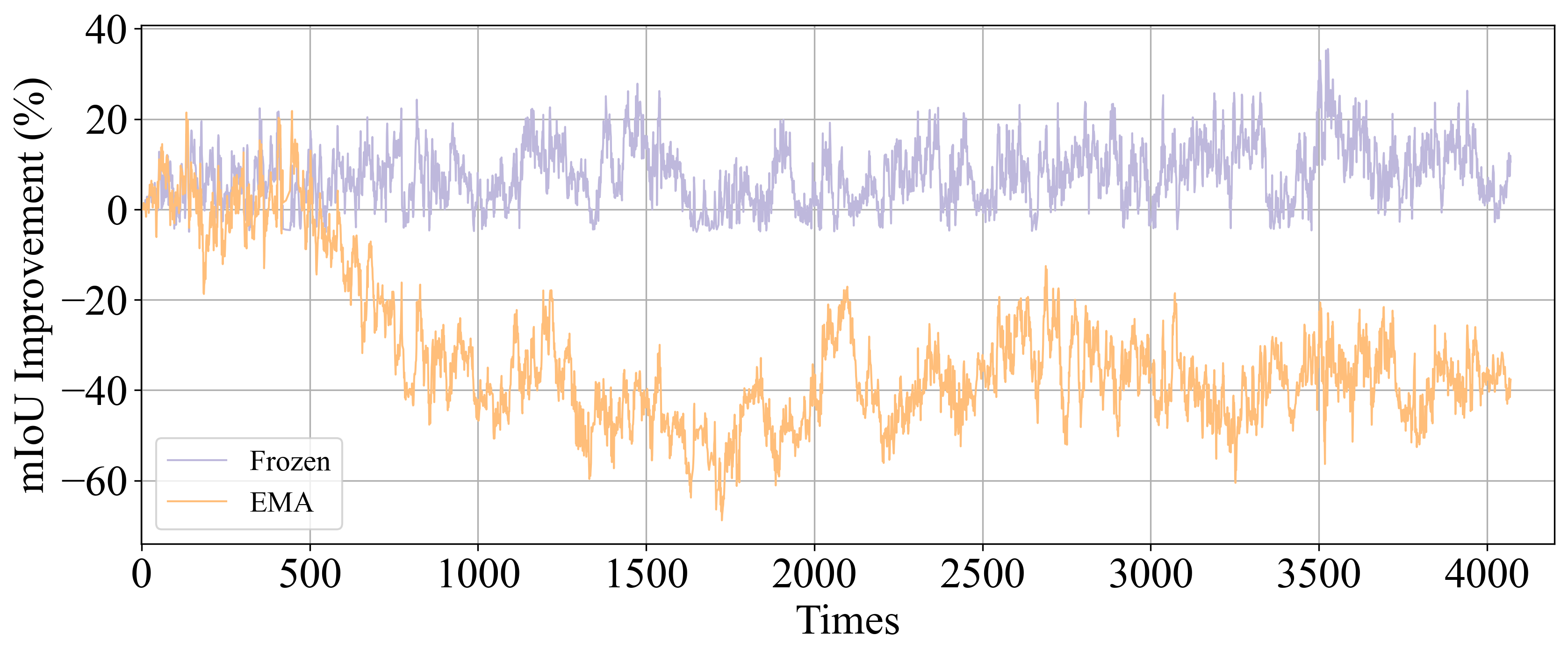}
\caption{mIoU improvement (\%) comparison over time between the frozen source and EMA model.}
\label{fig:frozen_and_ema}
\end{minipage}
\begin{minipage}[t]{0.48\textwidth}
\centering
\includegraphics[width=0.99\textwidth]{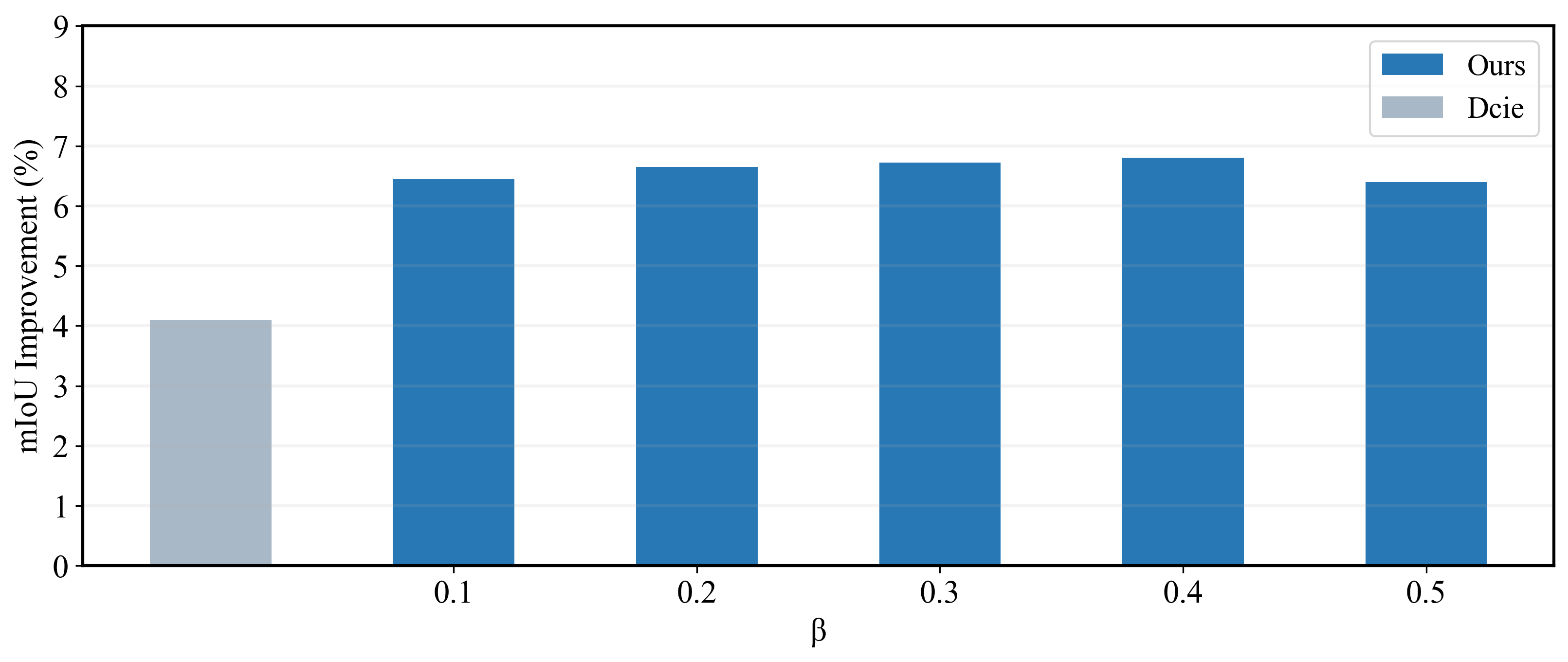}
\caption{Hyper-parameter sensitivity analysis of $\beta$ on SynLiDAR to SemanticKITTI. Dice is the basic dice loss.}
\label{fig:hyper_param_supple}
\end{minipage}
\end{figure}

\begin{table*}[htbp]
\centering
\caption{mIoU improvement (\%) in SynLiDAR to SemanticKITTI test-time adaptation with a different time window $w$. }
\addtolength{\tabcolsep}{+2.0pt}
\resizebox{0.99\textwidth}{!}{
\begin{tabular}{lccccccccc}
\toprule
\textbf{Model} &\textbf{$w$} & \textbf{Vehicle} & \textbf{Pedestrian} & \textbf{Road} & \textbf{Sidewalk} & \textbf{Terrain} & \textbf{Manmade} & \textbf{Vegetation} & \textbf{Avg} \\
\midrule
Source      & -                    & 59.80            & 14.20               & 34.90         & 53.50             & 31.00            & 37.40            & 50.50               & 40.19     \\
Upper Bound & -                      & +21.32           & +8.09               & +11.51        & +28.13            & +40.46           & +33.67           & +30.63              & +24.83      \\
\midrule
Ours        & 1                     & +9.07    & +5.34       & -1.35 & +1.46     & +1.29    & +4.36    & +5.72       & +3.70 \\
Ours        & 2                     & +11.95   & +5.39       & +0.76  & +3.65     & +2.55    & +4.89    & +5.33       & +4.93 \\
Ours        & 3                     & +14.79   & +5.81       & +0.81  & +3.63     & +4.07    & +4.66    & +5.32       & +5.58 \\
Ours        & 4                     & +12.74   & +5.25       & +2.06  & +4.82     & +4.23    & +6.21    & +6.53       & +5.98 \\
Ours        & 5                     & +14.76   & +5.66       & +1.83  & +5.43     & +7.33    & +5.64    & +6.40       & +6.72 \\
Ours        & 6                     & +12.88   & +5.96       & +1.32  & +4.64     & +5.71    & +7.87    & +8.82       & +6.74 \\
Ours        & 7                     & +13.34   & +6.49       & +2.54  & +5.47     & +6.13    & +8.90    & +10.28      & +7.59 \\
Ours        & 8                     & +13.37   & +4.97       & +3.56  & +5.25     & +6.82    & +10.03   & +11.60      & +7.94 \\
Ours        & 9                     & +13.89   & +4.34       & +1.81  & +4.75     & +5.34    & +9.69    & +11.01      & +7.26 \\
Ours        & 10                    & +12.74   & +5.85       & +2.52  & +4.80     & +5.04    & +8.42    & +10.05      & +7.06 \\
\bottomrule
\end{tabular}
}
\label{table:w}
\end{table*}

\subsection{More Experimental Analysis}
We conduct more experimental analysis to show the contribution of each component. Unless otherwise specified, the ablation analysis is based on SynLiDAR to SemanticKITTI. 

\noindent \textbf{Frozen Model or EMA:} As mentioned before, our Local Geometry Learning (LGL) is based on the frozen source model, rather than the Exponential Moving Average (EMA). To validate the reasonableness of our settings, we conduct experiments by replacing the frozen source model with EMA, formulated as:
\begin{equation}
\hat{F}_S^t=\delta \hat{F}_S^{t-1}+(1-\delta)F_T^t
\end{equation}
where $\hat{F}_S^0={F}_S^0$, $\hat{{F}}_S^t$ and $\hat{{F}}_S^{t-1}$ mean the current $t$-th model and last $(t-1)$-th model in iterations, and $\delta$ is a hyper-parameter that controls the updating speed. 
As shown in Fig. \ref{fig:frozen_and_ema}, EMA's performance is poorer than the frozen source model, with a mIoU of -30.94 \%.
We speculate that the underlying reason lies in that EMA is $\delta ^t {F}_S^{0}$ over the $t$ training step and ${F}_S^{0}$ is the pre-trained source model. Since $\delta ^t \to 0$ when $0 < \delta < 1$, the fine-tuned model with EMA will almost forget the knowledge of the pre-trained source model. Moreover, training of TTA is highly sensitive to parameters~\cite{wang2020tent}, which may result in a representational collapse during the EMA training process. We further illustrated the mIoU improvement comparison between the frozen source model and EMA over time through Fig. \ref{fig:frozen_and_ema}, which clearly shows the collapse at a sometime step.

\begin{table}[tp]
\begin{minipage}{.47\linewidth}
  \centering
  \caption{Ablation study of Local Geometry Learning. \textbf{NA}: Neighboring Aggregation. \textbf{PC}: Prediction Certainty. \textbf{GP}: Geometric Purity. \textbf{PA}: Pseudo-label Accuracy.}
  \addtolength{\tabcolsep}{+5.0pt}
  \resizebox{1.0\textwidth}{!}{
        \begin{tabular}{c|ccc|cc}
        \toprule
        ID     & NA        & PC        & GP        & PA (\%)       & mIoU (\%)      \\
        \midrule
        baseline &   & \checkmark           &              & 78.36 &+6.06\\ 
        \midrule
        I & \checkmark & \checkmark           &            &  79.95          & +6.42     \\
        II      & \checkmark & \checkmark & \checkmark           &80.05            &+6.72\\
        \bottomrule
        \end{tabular}
    }
  \label{table:LGL}%
\end{minipage}
\ \ \ 
\begin{minipage}{.47\linewidth}
  \centering
  \caption{mIoU improvement (\%) results on different Dice loss. Dice is basic dice loss without the smooth label. Soft Dice means dice loss with smooth label of fixed $\beta$. Soft Dice (Adaptive) means dice loss with adaptive $\beta$.}
  \addtolength{\tabcolsep}{+1.0pt}
  \resizebox{1.0\textwidth}{!}{
    \begin{tabular}{lccc}
    \toprule
             & Dice & Soft Dice & Soft Dice(Adaptive) \\
    \midrule
    mIoU(\%) &4.10           & 5.99                    & 6.72          \\            
    \bottomrule
    \end{tabular}
    }
  \label{table:dice_loss}%
\end{minipage}
\end{table}%

\noindent \textbf{Local Geometry Learning:}  We ablate components of Local Geometry Learning (LGL) in Table \ref{table:LGL}. Results show that the introduction of local geometric information through neighboring aggregation can significantly improve the accuracy of pseudo-label and result in improved performance compared to the baseline method based only on prediction certainty. Furthermore, although the introduction of geometric purity contributes less to the pseudo-label accuracy, it plays a substantial role in enhancing overall performance. As shown in the previous Fig. 5,  we speculate that this may be attributed to the fact that LGL with geometric purity will be more inclined to select points with features that are more easily distinguishable within the object, rather than points with features that are prone to confusion at the boundary. The preference for selecting points within the object is advantageous for the subsequent Global Geometry Fine-tuning (GGF) in introducing global (object-level) geometric information.


\noindent \textbf{Confidence Weight:} We experimented with different confidence weights in temporal geometry regularization, including confidence based on the target model's classification head and confidence based on global prototypes. Despite the introduction of continuously growing target data information, neither approach yields satisfactory results, with the former showing a mIoU improvement of +5.79, and the latter showing an improvement of +5.90.  These results are inferior to the confidences obtained through LGL. We hypothesize that there are two potential reasons for this. Firstly, the confidence of LGL includes local geometric information which contributes to training. Secondly, due to the highly sensitive nature of TTA tasks, fluctuating confidences may lead to sub-optimal performance.

\noindent \textbf{Soft Dice Loss:} Our adaptive soft dice loss can effectively reduce the negative transfer and improve the stability of training. As shown in Table~\ref{table:dice_loss}, we compare the model with basic dice loss and fixed soft dice loss. It shows that simply using the basic dice loss worsens the performance from +6.72\% to +4.10\%. 
We also conduct additional experiments on different loss functions. As shown in Table~\ref{table:diff_loss}, SoftDICE loss outperforms other loss functions, particularly for small-scale categories such as pedestrians. This improvement contributes to better overall segmentation results.

\noindent \textbf{Hyper-parameter Sensitivity:} We study the effect of different time window length $w$ and smoothing parameter $\beta$. Table~\ref{table:w} indicates that the performance is optimal when the value of $w$ is set to 8,  and performance degrades when $w$ is too small. Despite the significant performance improvement when $w=8$ compared to $w=5$, in order to maintain a fair comparison with~\cite{saltori2022gipso} and to facilitate proportional scaling $w$ on nuScenes, we still choose $w=5$. 
In Fig. \ref{fig:hyper_param_supple}, we present the mIoU improvement for $\beta$. It is easy to find that results around the selected parameter $\beta = 0.3$  are stable, and much better than the dice loss. 

\noindent \textcolor{black}{\textbf{Compared to domain generalization (DG) methods:} We last present a comparison between the DG method~\cite{kim2023single} and TTA methods. DGLSS obtains {$\mathbf{+3.0\%}$} mIoU on Synth4D(64) to SemanticKITTI, which is slightly less effective compared to TTA methods. It is important to note that DG is not direct competition with TTA, rather, it can be harnessed to enhance performance. By integrating HGL with DGLSS, we observe an improvement from {$\mathbf{+3.0\%}$} to {$\mathbf{+7.5\%}$} mIoU.}

\begin{table}[t]
\centering
\addtolength{\tabcolsep}{+18.0pt}
\caption{{mIoU(\%) and Pedestrians IoU(\%) improvement of different loss.}}
\resizebox{0.80\textwidth}{!}{
\begin{tabular}{l|ccccc}
\toprule
       & CE    & SoftCE & SCE   & DICE  & SoftDICE \\
\hline
mIoU      & +2.77 & +4.56  & +3.80 & +4.10 & \textbf{+6.72}    \\
Ped. IoU & -4.68 & +1.03  & +0.56 & -5.67 & \textbf{+5.66}    \\
\bottomrule
\end{tabular}
}
\label{table:diff_loss}
\end{table}

\begin{figure*}[th]
    \centering
    \includegraphics[width=0.99\textwidth]{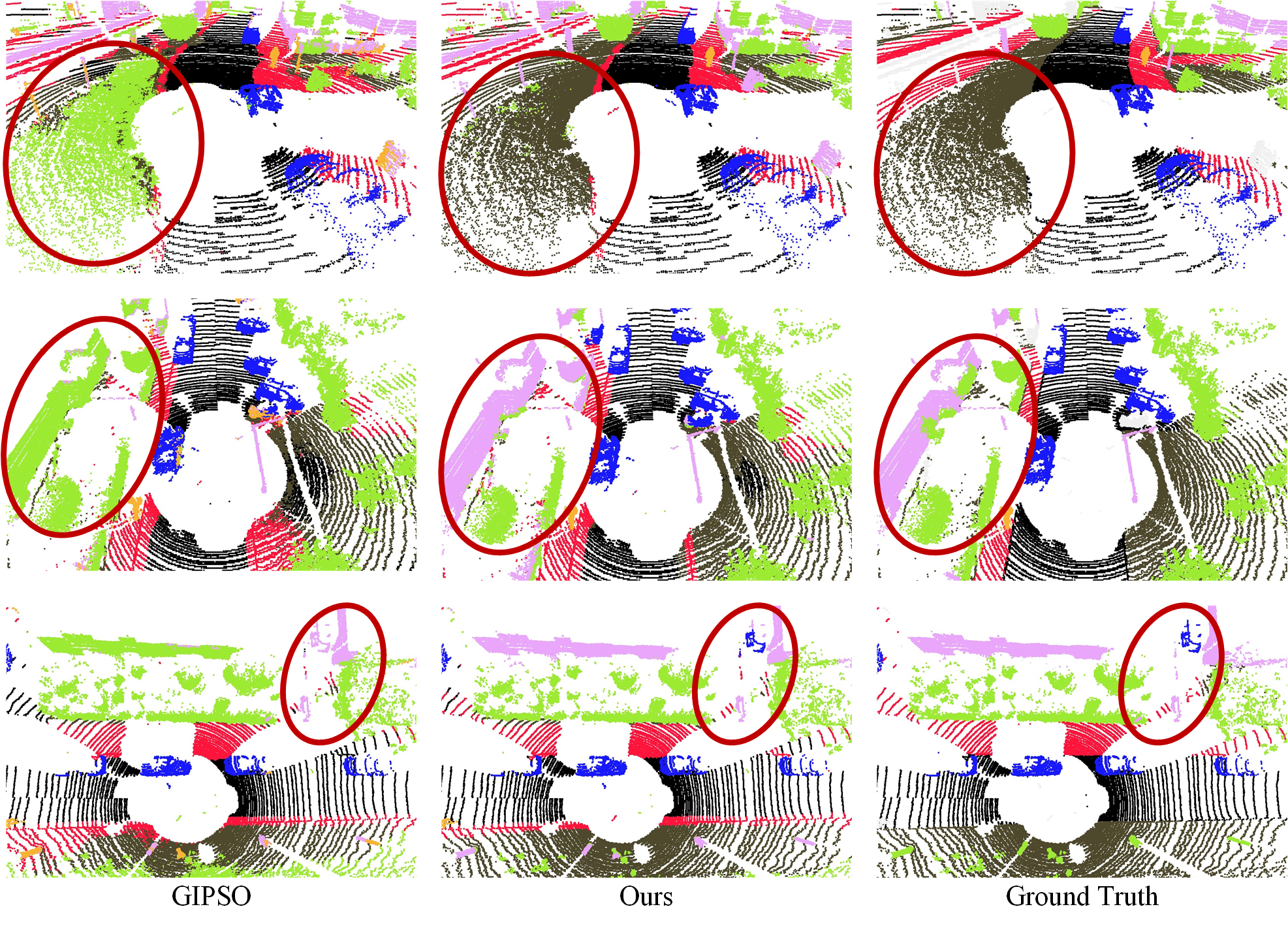}
    \caption{Visualization of adaptation result for the task Synth4D to SemanticKITTI.}
    \label{fig:vis2}
\end{figure*}

\begin{figure*}[th]
    \centering
    \includegraphics[width=0.99\textwidth]{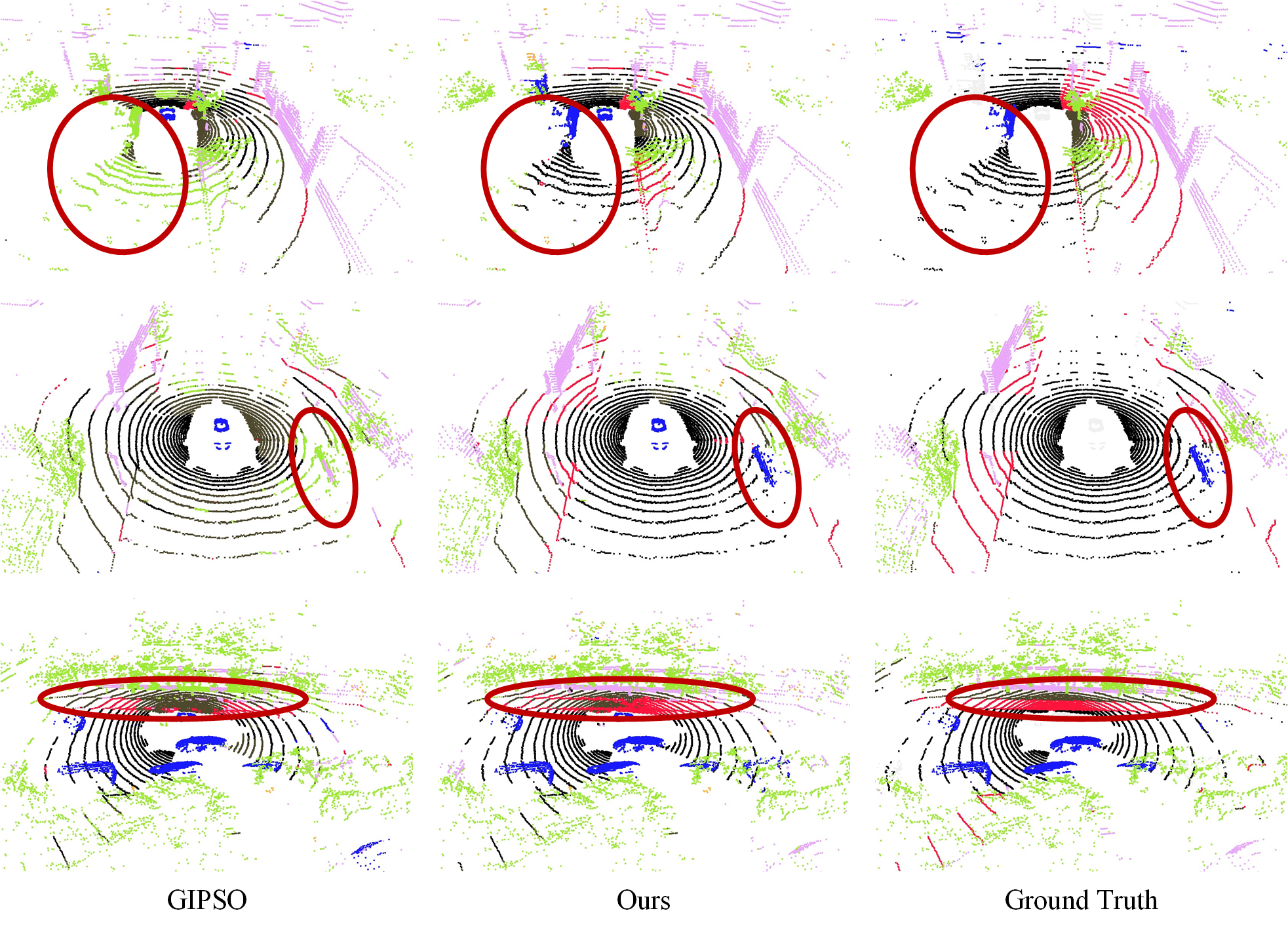}
    \caption{Visualization of adaptation result for the task Synth4D to nuScenes.}
    \label{fig:vis3}
\end{figure*}

\section{Visualization results}
In Fig. \ref{fig:vis2} and Fig. \ref{fig:vis3}, we report additional adaptation visualization results of the GIPSO, HGL, and ground truth in Synth4D to SemanticKITTI, and Synth4D to nuScenes. We observe that the results of our method are generally superior to the GIPSO.

\section{Class mapping} Due to the different annotations across datasets, we follow GIPSO~\cite{saltori2022gipso} to remap semantic categories of SynLiDAR~\cite{xiao2021synlidar}, SemanticKITTI~\cite{behley2019semantickitti} and nuScenes~\cite{caesar2020nuscenes} into seven standardized categories in Synth4D~\cite{saltori2022gipso} to ensure consistent category definitions. Table \ref{tab:kitti2synth4d}-\ref{tab:synlidarsynth4d} show the class mapping from SemanticKITTI to Synth4D, nuScenes to Synth4D, and SynLiDAR to Synth4D respectively. Categories that do not intersect with other datasets are considered unlabeled.

\begin{table}[!h]
    \centering
    \caption{Class mapping from SemanticKITTI to Synth4D.}
    \resizebox{0.60\textwidth}{!}{
    \begin{tabular}{cccc}
        \toprule
        SemanticKITTI-ID & SemanticKITTI-Name &  Synth4D-Name & Synth4D-ID\\
        \midrule
        0 & unlabelled & unlabelled & 0 \\
        1 &	car & vehicle & 1 \\
        2 &	bicycle & unlabelled & 0 \\
        3 &	motorcycle & unlabelled & 0 \\
        4 &	truck &	unlabelled & 0 \\
        5 &	other-vehicle &	unlabelled & 0 \\
        6 &	person & pedestrian & 2 \\
        7 &	bicyclist &	unlabelled & 0 \\
        8 &	motorcyclist & unlabelled & 0 \\
        9 &	road & road & 3 \\
        10 & parking& road & 3 \\
        11 & sidewalk & sidewalk & 4 \\
        12 & other-ground & unlabelled & 0 \\
        13 & building& manmade & 6 \\
        14 & fence & manmade & 6 \\
        15 & vegetation& vegetation & 7 \\
        16 & trunk & vegetation & 7 \\
        17 & terrain & terrain & 5 \\
        18 & pole &	manmade & 6 \\
        19 & traffic-sign & manmade & 6 \\
        \bottomrule
    \end{tabular}
    }
    \label{tab:kitti2synth4d}
\end{table}

\begin{table}[t]
    \centering
    \caption{Class mapping from nuScenes to Synth4D.}
    \resizebox{0.60\textwidth}{!}{
    \begin{tabular}{cccc}
        \toprule
        nuScenes-ID & nuScenes-Name &  Synth4D-Name & Synth4D-ID\\
        \midrule
        0 & unlabelled & unlabelled & 0 \\
        1 & barrier & unlabelled & 0 \\
        2 & bicycle & unlabelled & 0 \\
        3 & bus & unlabelled & 0 \\
        4 & car & vehicle & 1\\
        5 & construction-vehicle & unlabelled & 0 \\
        6 & motorcycle & unlabelled & 0\\
        7 & pedestrian & pedestrian & 2 \\
        8 & traffic-cone & unlabelled & 0 \\
        9 & trailer & unlabelled & 0 \\
        10 & truck & unlabelled & 0\\
        11 & driveable-surface & road & 3 \\
        12 & other-flat & unlabelled & 0 \\
        13 & sidewalk & sidewalk & 4\\
        14 & terrain & terrain & 5\\
        15 & manmade & manmade & 6 \\
        16 & vegetation & vegetation & 7\\
        \bottomrule
    \end{tabular}
    }
    \label{tab:nusc2synth4d}
\end{table}

\begin{table}[t]
    \centering
    \caption{Class mapping from SynLiDAR to Synth4D.}
    \resizebox{0.60\textwidth}{!}{
    \begin{tabular}{cccc}
        \toprule
        SynliDAR-ID & SynLiDAR-Name &  Synth4D-Name & Synth4D-ID\\
        \midrule
        0 & unlabelled & unlabelled & 0\\
        1 & car & vehicle & 1\\
        2 & pickup & vehicle & 1\\
        3 & truck & unlabelled & 0\\
        4 & bus & unlabelled & 0\\
        5 & bicycle & unlabelled & 0\\
        6 & motorcycle & unlabelled & 0\\
        7 & other-vehicle & unlabelled & 0\\
        8 & road & road & 3\\
        9 & sidewalk & sidewalk & 4\\
        10 & parking & road & 3\\
        11 & other-ground & unlabelled & 0\\
        12 & female & pedestrian & 2\\
        13 & male & pedestrian & 2\\
        14 & kid & pedestrian & 2\\
        15 & crowd & pedestrian & 2\\
        16 & bicyclist & unlabelled & 0\\
        17 & motorcyclist & unlabelled & 0\\
        18 & building & manmade & 6\\
        19 & other-structure & unlabelled & 0\\
        20 & vegetation & vegetation & 7\\
        21 & trunk & vegetation & 7\\
        22 & terrain & terrain & 5 \\
        23 & traffic-sign & manmade & 6\\
        24 & pole & manmade & 6\\
        25 & traffic-cone & unlabelled & 0\\
        26 & fence & manmade & 6\\
        27 & garbage-can & unlabelled & 0\\
        28 & electric-box & unlabelled & 0\\
        29 & table & unlabelled & 0\\
        30 & chair & unlabelled & 0\\
        31 & bench & unlabelled & 0\\
        32 & other-object & unlabelled & 0\\
        \bottomrule
    \end{tabular}
    }
    \label{tab:synlidarsynth4d}
\end{table}

%
%
\bibliographystyle{splncs04}
\bibliography{main}